\documentclass{SCIS2019}
\usepackage{multirow}
\usepackage{wrapfig}
\newcommand{\tabincell}[2]{\begin{tabular}{@{}#1@{}}#2\end{tabular}}
\usepackage[numbers,sort&compress]{natbib}
\begin{document}
%\oa
%%%%%%%%%%%%%%%%%%%%%%%%%%%%%%%%%%%%%%%%%%%%%%%%%%%%%%%
%%% Authors do not modify the information below
%%% 作者不需要修改此处信息
\ArticleType{RESEARCH PAPER}
%\SpecialTopic{}
\Year{2023}
\Month{}
\Vol{}
\No{}
\DOI{}
\ArtNo{}
\ReceiveDate{}
\ReviseDate{}
\AcceptDate{}
\OnlineDate{}
%%%%%%%%%%%%%%%%%%%%%%%%%%%%%%%%%%%%%%%%%%%%%%%%%%%%%%%

%%% title: 标题
%%%   \title{title}{title for citation}
\title{EmotionIC: emotional inertia and contagion-driven dependency modeling for emotion recognition in conversation}{Received 23 March 2023/Revised 28 August 2023/Revised 16 October 2023/Accepted 10 November 2023}

%%% Corresponding author: 通信作者
%%%   \author[number]{Full name}{{email@xxx.com}}
%%% General author: 一般作者
%%%   \author[number]{Full name}{}
\author[]{Yingjian LIU\textsuperscript{$134\dagger$}}{}
\author[]{Jiang LI\textsuperscript{$1234*\dagger$}}{lijfrank@hust.edu.cn (Li J)}
\author[]{Xiaoping WANG\textsuperscript{134$*$}}{wangxiaoping@hust.edu.cn (Wang X P)}
\author[]{Zhigang ZENG\textsuperscript{134}}{}
% \author[134]{Yingjian LIU\textsuperscript{$\dagger$}}{}
% \author[1234]{Jiang LI\textsuperscript{$*\dagger$}}{lijfrank@hust.edu.cn (Li J)}
% \author[134]{Xiaoping WANG\textsuperscript{$*$}}{wangxiaoping@hust.edu.cn (Wang X P)}
% \author[134]{Zhigang ZENG}{}

%%% Author information for page head. 页眉中的作者信息
\AuthorMark{Liu Y J, Li J}

%%% Authors for citation. 首页引用中的作者信息
\AuthorCitation{Liu Y J, Li J, Wang X P, et al}

%%% Authors' contribution. 同等贡献
\contributions{Yingjian LIU and Jiang LI have the same contribution to this work.}

%%% Address. 地址
%%%   \address[number]{Affiliation, City {\rm Postcode}, Country}
% \address[]{School of Artificial Intelligence and Automation, Huazhong University of Science and Technology, Wuhan 430074, China\\
% Institute of Artificial Intelligence, Huazhong University of Science and Technology (HUST), Wuhan 430074, China\\
% Key Laboratory of Image Processing and Intelligent Control of Education Ministry of China, Wuhan 430074, China\\
% Hubei Key Laboratory of Brain-inspired Intelligent Systems, Wuhan 430074, China}
\address[]{\textsuperscript{1}School of Artificial Intelligence and Automation, Huazhong University of Science and Technology, \\Wuhan {\rm 430074}, China\\
\textsuperscript{2}Institute of Artificial Intelligence, Huazhong University of Science and Technology, Wuhan {\rm 430074}, China\\
\textsuperscript{4}Hubei Key Laboratory of Brain-inspired Intelligent Systems, Huazhong University of Science and Technology, \\Wuhan {\rm 430074}, China\\
\textsuperscript{3}Key Laboratory of Image Processing and Intelligent Control (Huazhong University of Science and Technology), \\Ministry of Education, Wuhan {\rm 430074}, China}

%%% Abstract. 摘要
\abstract{Emotion Recognition in Conversation (ERC) has attracted growing attention in recent years as a result of the advancement and implementation of human-computer interface technologies. In this paper, we propose an emotional inertia and contagion-driven dependency modeling approach (EmotionIC) for ERC task. Our EmotionIC consists of three main components, i.e., Identity Masked Multi-Head Attention (IMMHA), Dialogue-based Gated Recurrent Unit (DiaGRU), and Skip-chain Conditional Random Field (SkipCRF). Compared to previous ERC models, EmotionIC can model a conversation more thoroughly at both the feature-extraction and classification levels. The proposed model attempts to integrate the advantages of attention- and recurrence-based methods at the feature-extraction level. Specifically, IMMHA is applied to capture identity-based global contextual dependencies, while DiaGRU is utilized to extract speaker- and temporal-aware local contextual information. At the classification level, SkipCRF can explicitly mine complex emotional flows from higher-order neighboring utterances in the conversation. Experimental results show that our method can significantly outperform the state-of-the-art models on four benchmark datasets. The ablation studies confirm that our modules can effectively model emotional inertia and contagion.}

%%% Keywords. 关键词

\maketitle

%%%%%%%%%%%%%%%%%%%%%%%%%%%%%%%%%%%%%%%%%%%%%%%%%%%%%%%
%%% The main text. 正文部分
%%%%%%%%%%%%%%%%%%%%%%%%%%%%%%%%%%%%%%%%%%%%%%%%%%%%%%%
\section{Introduction}
Emotion Recognition in Conversation (ERC) is one of the most focusing research fields in Natural Language Processing (NLP), which aims to identify the emotion of each utterance in a conversation. This task has recently received considerable attention from NLP researchers due to its potential applications in multiple domains such as opinion mining in social media~\cite{8267597,habimana2020sentiment}, empathic dialogue system construction~\cite{majumder2019dialoguernn, lin2019moel}, and smart home systems~\cite{young2018augmenting,khan2022human}. Emotions are often reflected in interpersonal interactions, and analyzing the emotions of a single utterance out of the conversational context may lead to ambiguity~\cite{zhou2018emotional}. Therefore, ERC incorporating conversational context information significantly contributes to model performance. %rashkin2018towards,

The effective use of contextual information in dialogues lies at the heart of ERC~\cite{poria2019emotion}. There are numerous efforts have been developed to encode the contextual information in the dialogue, including graph-based methods~\cite{ghosal2019dialoguegcn,shen-etal-2021-directed,li2022graphcfc}, recurrence-based methods~\cite{majumder2019dialoguernn, jiao2019higru,Ghosal2020}, and attention-based methods~\cite{zhong2019knowledge,zhu2021kat,li2022ga2mif}. Li et al.~\cite{li2021past} proposed a psychological-knowledge-aware interaction graph, which established four relations in a local connectivity graph to simulate the psychological state of the speaker. Hu et al.~\cite{hu2021dialoguecrn} designed a multi-turn reasoning module based on the Recurrent Neural Network (RNN), which iteratively performed the intuitive retrieval process and conscious reasoning process to extract and integrate emotional cues from a cognitive perspective. Zhu et al.~\cite{zhu2021kat} proposed a topic-driven and knowledge-aware Transformer model that incorporated topic representation and the commonsense knowledge from ATOMIC for emotion detection in dialogues. 

However, recurrence-based methods tend to use only relatively limited information from recent utterances to update the state of the current utterance, which makes them difficult to achieve satisfying performance; graph- and attention-based approaches diminish the importance of neighboring utterances because of global relevance, resulting in the loss of temporal sequential information in the conversation. According to the above analysis, a better way to implement ERC is to combine the strengths of attention- and recurrence-based models. The neighboring utterances tend to contain information about emotional inertia and contagion, and this information can diminish over time. Despite the weak influence of long-distance contexts on the current utterance, they imply abundant global information that can assist in classifying utterances without clear emotions. Thus, we model contexts of the current utterance by utilizing the speaker identity-based Multi-Head Attention (MHA)~\cite{vaswani2017attention} and Gated Recurrent Unit (GRU)~\cite{chung2014empirical} to extract global and local information, respectively.

During the conversation, a speaker's emotion is influenced by his/her own or others' historical emotions, indicating that there are significant dependencies between emotions in the conversation. Existing ERC models focus on contextual modeling at the feature-extraction level and rarely mine the emotional flows in the conversation at the classification level. To achieve this purpose, we draw on the effectiveness of Conditional Random Field (CRF) for modeling sequential dependencies to explicitly model emotional interactions in the conversation. Not only that, but in order to simulate the complex emotional propagations in the conversation, we introduce skip connections in CRF to capture emotion influence from higher-order neighboring utterances (i.e., indirect neighbors).

\begin{figure}[htbp]
	\centering
	\includegraphics[height=2.0in]{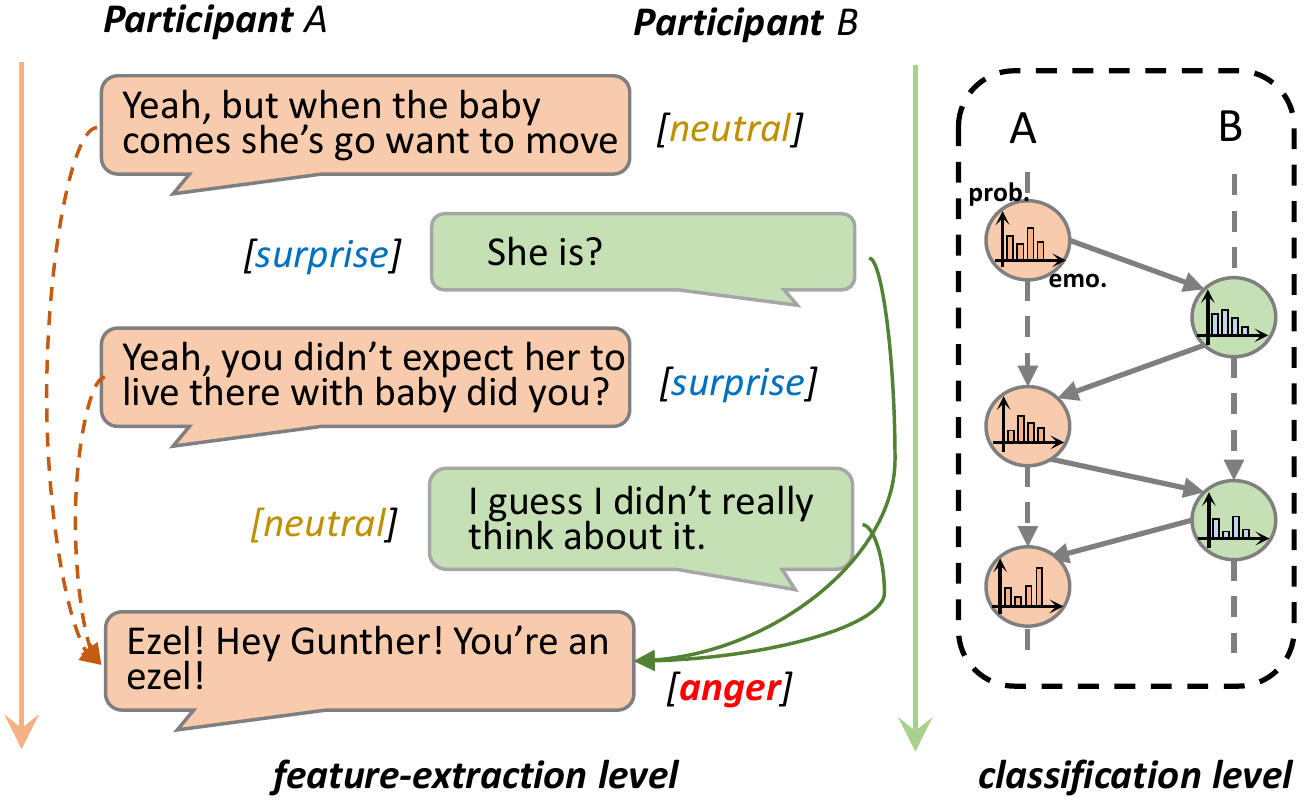}
	\caption{Example of contextual dependency modeling at both the feature-extraction and classification levels. The dashed and solid line represent the intra- and inter-speaker information transmissions, respectively.}
	\label{fig:example}
\end{figure}
In summary, we propose an \textbf{Emotion}al \textbf{I}nertia and \textbf{C}ontagion-driven dependency modeling approach (EmotionIC) in this paper, which can adequately model contexts from the feature-extraction and classification levels. Figure~\ref{fig:example} is an example of modeling context dependency at both the feature-extraction and classification levels. At the feature-extraction level, we design an Identity Masked MHA (IMMHA) to capture intra- and inter-speaker dependencies in the global contexts from two features subspaces, respectively; to further refine the contextual dependencies, we devise a speaker- and position-aware Dialogue GRU (DiaGRU). DiaGRU introduces the emotional tendencies of the current speaker and interlocutor at the previous moment into the single GRU cell. At the classification level, by introducing skip connections in CRF, we craft a novel structure called Skip-chain CRF (SkipCRF) to explicitly capture the emotional flows in the conversation. SkipCRF takes into account higher-order contextual dependencies from intra- and inter-speaker, and can model the complex emotional interactions of different participants in the conversation. It is worth noting that instead of using additional softmax layer, we directly utilize CRF for final emotion classification. Experimental results demonstrate the superiority of our model compared with state-of-the-art models, and several studies are conducted to illustrate the effectiveness of each module of EmotionIC.
To put it briefly, our main contributions are summarized as follows:
\begin{itemize}
	\item We propose a novel model, EmotionIC, for the ERC task. EmotionIC models a conversation thoroughly at both the feature-extraction and classification levels and is mainly composed of IMMHA, DiaGRU, and SkipCRF.
	\item At the feature-extraction level, we combine the strengths of attention- and recurrence-based models. IMMHA extracts identity-based global contextual information, while DiaGRU captures participant- and temporal-aware local contextual information.
	\item At the classification level, SkipCRF can extract complex emotional flows from higher-order neighboring utterances in the conversation while accomplishing final emotion classification.
	\item We perform extensive experiments on the IEMOCAP, DailyDialog, MELD, and EmoryNLP datasets and obtain the most advanced performance, which demonstrates the superiority of the proposed method.
\end{itemize}

\section{Related work}\label{sec:related_work}
Emotion recognition is an interdisciplinary field of research, with contributions from different fields such as natural language processing, computer vision, and psychological cognitive science~\cite{picard2010affective}. In this section, we mainly introduce the related works of emotion recognition in conversation and conditional random field. Moreover, we briefly introduce the applications of CRF in ERC tasks.

\subsection{Emotion recognition in conversation}
Distinct from traditional emotion recognition which treats emotion as a static state, Emotion Recognition in Conversation (ERC) takes full consideration of emotion to be dynamic and flow between speaker interactions. Hazarika et al.~\cite{hazarika2018icon} proposed a model based on Long- and Short-Term Memory (LSTM) to enable current utterance to capture contextual information in historical conversations. CMN~\cite{hazarika2018conversational} employed a skip attention mechanism to merge contextual information in a historical conversation. Jiao et al.~\cite{jiao2019higru} proposed a hierarchical GRU to address the difficulty of capturing long-distance contextual information effectively. By distinguishing specific speakers, DialogueRNN~\cite{majumder2019dialoguernn} modeled emotions dynamically based on the current speaker, contextual content, and emotional state. Zhong et al.~\cite{zhong2019knowledge} proposed Knowledge-Enriched Transformer, which dynamically exploited external commonsense knowledge through hierarchical self-attention and context-aware graph attention. By building directed graphical structures over the input utterance sequences with speaker information, DialogueGCN~\cite{ghosal2019dialoguegcn} applied graph convolution network to construct intra- and inter-dependencies among distant utterances. COSMIC~\cite{Ghosal2020} combined different commonsense knowledge and learned the interaction between the interlocutors in the dialogue. DialogXL~\cite{shen2021dialogxl} modified the memory block in XLNet~\cite{yang2019xlnet} to store longer historical contexts and conversation-aware self-attention to handle multi-party structures. Wang et al.~\cite{wang2020relational} proposed a relational graph attention network to encode the tree structure for sentiment prediction. DAG-ERC~\cite{shen-etal-2021-directed} treated the internal structure of dialogue as a directed acyclic graph, which intuitively model the way information flows between long and short distance contexts. Considering that utterances with similar semantics may have distinctive emotions under different contexts, CoG-BART~\cite{li2022contrast} adopted supervised contrastive learning to enhance the model's ability to handle context information. GAR-Net~\cite{xu2022gar} was an end-to-end graph attention reasoning network that took both word-level and utterance-level context into concern, aiming to emphasize the importance of contextual reasoning. Most prior efforts do not combine the strengths of global and local conceptual modeling and fail to explicitly consider self- and other-dependency based on emotional inertia and contagion.

\subsection{Conditional random field}
Conditional Random Fields (CRFs)~\cite{lafferty2001,MAL-013} are a class of probabilistic graphical modeling methods that aim to model the conditional distribution $\mathbf{P}{(Y|X)}$ by a set of observed variables $X=(X_1,X_2,\cdots,X_\mathcal{T})$, another set of unobserved variables $Y=(Y_1,Y_2,\cdots,Y_\mathcal{T})$, and the structural information among different variables. CRF relaxes the strong dependency assumptions in other Bayesian models based on directed graphical models and enables the establishment of higher-order dependency, which means that the result of CRF is closer to the real distribution of data~\cite{Li2022AComprehensiveReview}. CRF has recently attracted the interest of researchers in the ERC field~\cite{wang2020contextualized, song2022emotionflow, liang2021s+}, and these methods demonstrate the effectiveness of CRF for modeling emotion dependency at the classification level. CRF utilizes the potential function with clusters on the graph structure to define the conditional probability $\mathbf{P}{(Y|X)}$. Since there are primarily two types of clusters on labels in the linear-chain CRF frequently employed in ERC models, two forms of exponential potential functions are added as feature functions. Suppose the random variable $X$ and $Y$ take the values $\mathrm{x}=(x_1,x_2,\cdots,x_\mathcal{T})$ and $\mathrm{y}=(y_1,y_2,\cdots,y_\mathcal{T})$, respectively, $\mathbf{P}{(\mathrm{y}|\mathrm{x})}$ is defined formally as:
\begin{equation}
	\displaystyle
	\begin{aligned}
		\mathbf{P}{(\mathrm{y}|\mathrm{x})} &= \frac{1}{\mathbf{Z}(\mathrm{x})}\exp \big [\sum_{i,t}\lambda_i \bm{g}_i(y_{t-1},y_t,\mathrm{x},t)+\sum_{l,t}\mu_l \bm{f}_l(y_t,\mathrm{x},t) \big ]\\
		&=  \frac{1}{\mathbf{Z}(\mathrm{x})}\exp \big [\sum_{n,t}\omega_n \bm{F}_n(y_{t-1},y_t,\mathrm{x}) \big ],
	\end{aligned}
\label{eq:linercrf}
\end{equation}
where $\mathbf{Z}(\mathrm{x})$ represents the normalization factor; $\bm{F}_n(\cdot)$ is the feature function of linear-chain CRF, which consists of the local feature function $\bm{g}_i(\cdot)$ and the nodal feature function $\bm{f}_l(\cdot)$; $\bm{g}_i(\cdot)$ is defined on the context-connected edge of node $Y$, which means the state transition from $y_{t-1}$ to $y_t$; $\bm{f}_l(\cdot)$ is defined on node $Y$, which means the state of $y_t$; $\omega_n$ consists of $\lambda_i$ and $\mu_l$, which are learnable weights of the corresponding feature function. In the existing ERC methods for modeling this dependency, only the linear-chain CRF with first-order dependency is applied, i.e., only the dependency between neighbor tags are considered. This simple form is difficult to cope with complex interaction situations of different participants in dialogue scenarios. So we construct SkipCRF that introduces higher-order dependency through skip-chain connections to model emotional inertia and contagion at the classification level.

\section{Our approach}\label{sec:our_approach}
\begin{figure}[htbp]
	\centering
	\includegraphics[height=2.5in]{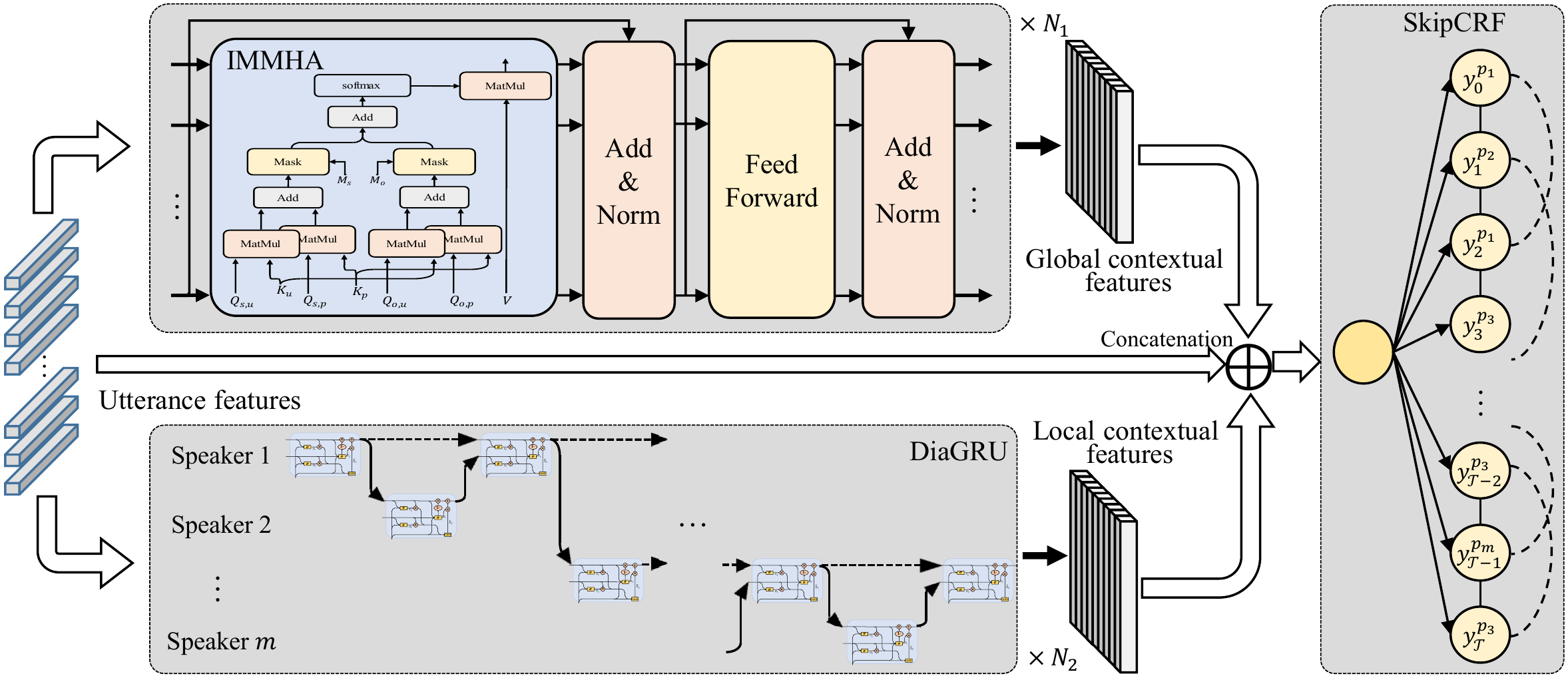}
	\caption{Architecture of our EmotionIC. Firstly, the global and local context dependencies are extracted through IMMHA and DiaGRU, respectively. Then, we concatenate global context features, local context features and utterance features. Note that combining utterance features that are not processed by IMMHA and DiaGRU is to prevent the original semantics of utterances with weak context dependencies from being obscured. Finally, the emotional flows in the conversation are captured at the classification level through SkipCRF to obtain the optimal emotion sequence.}
	\label{fig:overall}
\end{figure}

In this section, we will introduce the main components of our approach. First, we present the problem definitions and make certain transformations of the original problem according to the requirements of the proposed model. Then, we illustrate the architecture of our model as in Figure~\ref{fig:overall}, which contains three components: (1) Identity Masked Multi-head Attention (IMMHA), which captures global historical information from different participants; (2) Dialogue-based Gated Recurrent Unit (DiaGRU), which focuses on local intra- and inter-speaker dependencies of the current utterance; (3) Skip-chain Conditional Random Field (SkipCRF), which explicitly captures complex emotional flows at the classification level to obtain the optimal emotion sequence. In the following subsections, the key elements in these three components are described in detail.

\subsection{Preliminaries}
\begin{figure}[htbp]
	\centering
	\includegraphics[height=1.0in]{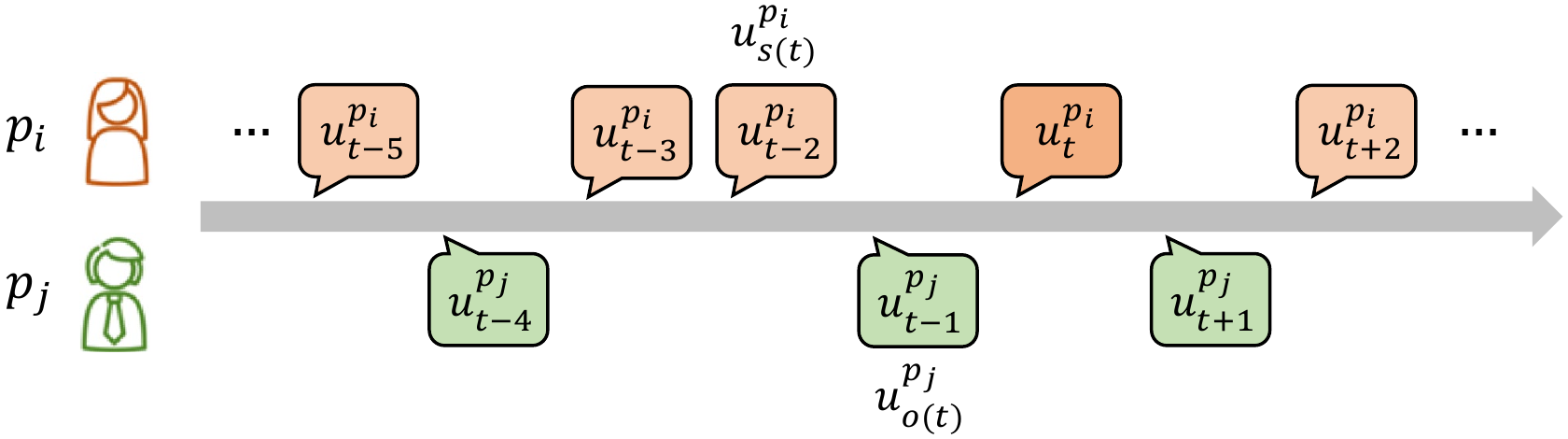}
	\caption{Illustration of utterance moment function. $u_{t-2}^{p_i}$ is the nearest and previous utterance uttered by the speaker $p_i$, so it can also be denoted by $u_{\bm{s}(t)}^{p_i}$. Similarly, $u_{t-1}^{p_j}$ can be also denoted by $u_{\bm{o}(t)}^{p_j}$.} 
	\label{fig:moment_func}
\end{figure}

\textbf{Problem definition.} The objective of the ERC task is to predict the emotion label $e_t$ corresponding to the $t$-th utterance $u_{t}^{p_i}$ in the conversation $C=(u_{1}^{p_1}, u_{2}^{p_2},\cdots,u_{\mathcal{T}}^{p_m})$ containing $m$ participants. Here, $e_t \in E$, and $E$ is the set of emotion labels in the dataset; $p_i\in \{p_1,p_2,\cdots,p_m\}$ is the speaker identity; $t$ describes the order of the utterance and also represents the moment corresponding to the current utterance; $\mathcal{T}$ represents the length of conversation $C$.
\\ \noindent 
\textbf{Speaker-specific utterance block.} Considering the continuous utterances of the same participant in the conversation as a speaker-specific utterance block, the original utterances sequence can be divided into multiple blocks sequence $(b_{1}^{p_1}, b_{2}^{p_2},\cdots, b_{\mathcal{B}}^{p_m})$, where $\mathcal{B}$ represents the length of utterance block, and $\mathcal{B}\le \mathcal{T}$. Note that a speaker-specific utterance block may contain one or more original utterances.
\\ \noindent 
\textbf{Utterance moment function.} To distinguish contextual dependencies from different participants, two utterance moment functions, i.e., $\bm{s}(\cdot)$ and $\bm{o}(\cdot)$, are defined in conversation. Here, $\bm{s}(\cdot)$ is utilized to output the moment of the previous utterance that belongs to the same speaker as the current utterance and is the nearest to that utterance; $\bm{o}(\cdot)$ is employed to output the moment of the previous utterance that is spoken by the interlocutor and is the nearest to the current utterance. As shown in Figure~\ref{fig:moment_func}, assuming the existence of the current utterance $u_{t}^{p_i}$, then $u_{\bm{s}(t)}^{p_i}$ represents the nearest and previous utterance uttered by the speaker $p_i$, and $u_{\bm{o}(t)}^{p_j}$ indicates the most recent utterance spoken by the interlocutor $p_j$.
\\ \noindent 
\textbf{Emotional inertia and contagion.} In addition to context dependency, speaker identity information is also shown to be critical to ERC~\cite{lili2020}. Research results in the field of psychology analyzed how emotions are transmitted in conversations~\cite{Hatfield1993}. Specifically, the transmission of emotion in interpersonal communication and dialogue is mainly driven by two factors: emotional inertia which means that a speaker in the conversation tends to maintain a particular emotional state (i.e., intra-speaker dependency or self-dependency), and emotional contagion which describes the emotional stimulation of other participants' utterances (i.e., inter-speaker dependency or other-dependency). In this paper, these two factors run through our methodology.
\\ \noindent
\textbf{Higher-order dependency.} \texttt{Higher-Order}~\cite{Li2022AComprehensiveReview,Yu2020Acomprehensivereview} is derived from concepts in the CRF domain. Refer to the definition of higher-order CRF, \texttt{Order} denotes the hop count of the current utterance's neighbor in our work. \texttt{Higher-Order} is a relative concept; in general, the order greater than or equal to 2 indicates \texttt{Higher-Order}. For instance, in conversation $(u_1,u_2,u_3,u_4,u_5)$, $u_2$ is the $3$rd-order neighbor (indirect/higher-order neighbor) of $u_5$, and $u_4$ is the $1$st-order neighbor (direct neighbor) of $u_5$. Accordingly, the meaning of higher-order dependencies is the dependencies of the current utterance on higher-order neighbors.

\subsection{Identity masked multi-head attention}
\begin{figure}[htbp]
	\centering
	\includegraphics[height=2.0in]{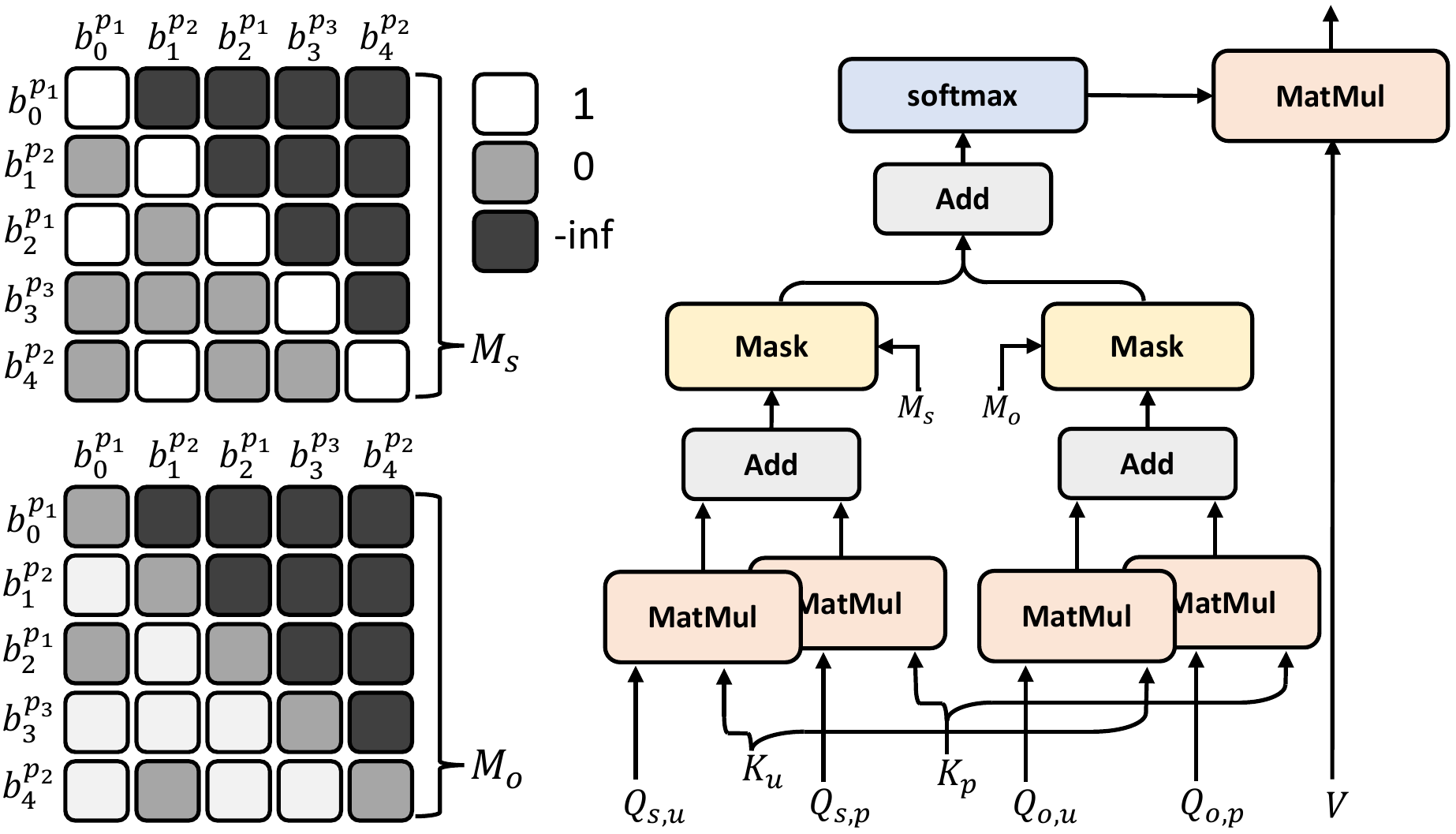}
	\caption{Network structure of IMMHA. $M_s$ and $M_o$ are two mask matrices that mask contextual dependencies from other participants and the current participant, respectively, as well as future information. $\mathbf{MatMul}$ indicates the matrix multiplication operation, and $\mathbf{Mask}$ is the masking operation.} 
	\label{fig:immha}
\end{figure}
Taking advantage of the ability of global context-awareness in conversation, attention mechanism is widely applied in many ERC models~\cite{zhu2021kat,zhong2019knowledge}. However, these approaches didn't explicitly encode the speaker identity information based on emotional inertia and contagion. In this subsection, we design a new Identity Masked Multi-Head Attention (IMMHA) that can effectively combine the identity information of participants to capture intra- and inter-speaker global dependencies in the conversation.

As shown on the left of Figure~\ref{fig:immha}, we add two mask matrices (i.e., $M_s$ and $M_o$) in IMMHA to capture intra- and inter-speaker contextual information, respectively. Here, $M_s$ is utilized to mask contextual dependencies from other participants (i.e., only the contextual information from the current participant is retained), and $M_o$ is implemented to mask contextual dependencies from the current participant (i.e., only the contextual information from other participants is maintained). Note that although we draw on multi-head self-attention~\cite{vaswani2017attention}, we treat utterances as inputs to IMMHA rather than tokens. For the capture of intra-speaker contextual information, we first matrix-multiply the utterance query matrix and utterance key matrix, and then mask the resulting result with the masked matrix $M_s$. In the actual implementation, we perform the untied absolute position encoding before the mask operation by referring to Ke et al.~\cite{KeG2020rethink}. Similar steps to those described above can be adopted to capture the inter-speaker contextual information. 
Finally, the captured intra- and inter-speaker contextual information is summed, which in turn is passed through the softmax layer to obtain the attention matrix of IMMHA. The whole process can be seen on the right of Figure~\ref{fig:immha} and can be described with the following formulation:
\begin{equation}
	{\rm{IM}}\text{-}{\rm{Attn}}= {\mathbf{softmax}}\big((Q_{s,u} K_u^\top + Q_{s,p} K_p^\top) \odot M_s
	+ (Q_{o,u} K_u^\top + Q_{o,p} K_p^\top) \odot  M_o\big),
	\end{equation}
where $Q_{s,u}$ and $Q_{o,u}$ represent the utterance query matrices, which are the results of mapping the utterances to different subspaces (i.e., self- and other-dependent subspaces) through the fully connected layers; $Q_{s,p}$ and $Q_{o,p}$ indicate the position query matrices; $K_u$ and $K_p$ are the utterance key matrices, which are the results of mapping the utterances to the same subspace through the fully connected layer; $\odot$ stands for the element-wise product.

To get the final output of IMMHA, the obtained attention matrix ${\rm{IM}}\text{-}{\rm{Attn}}$ is matrix multiplied with the utterance value matrix. In addition, inspired by the excellent architecture of Transformer~\cite{vaswani2017attention}, we pass the output of IMMHA through the residual, normalization, and feedforward layers, while adopting the multi-head setting.

\subsection{Dialogue-based gated recurrent unit}
\begin{figure}[htbp]
	\centering
	\captionsetup[subfigure]{}
	\subfloat[]{\includegraphics[height=1.2in]{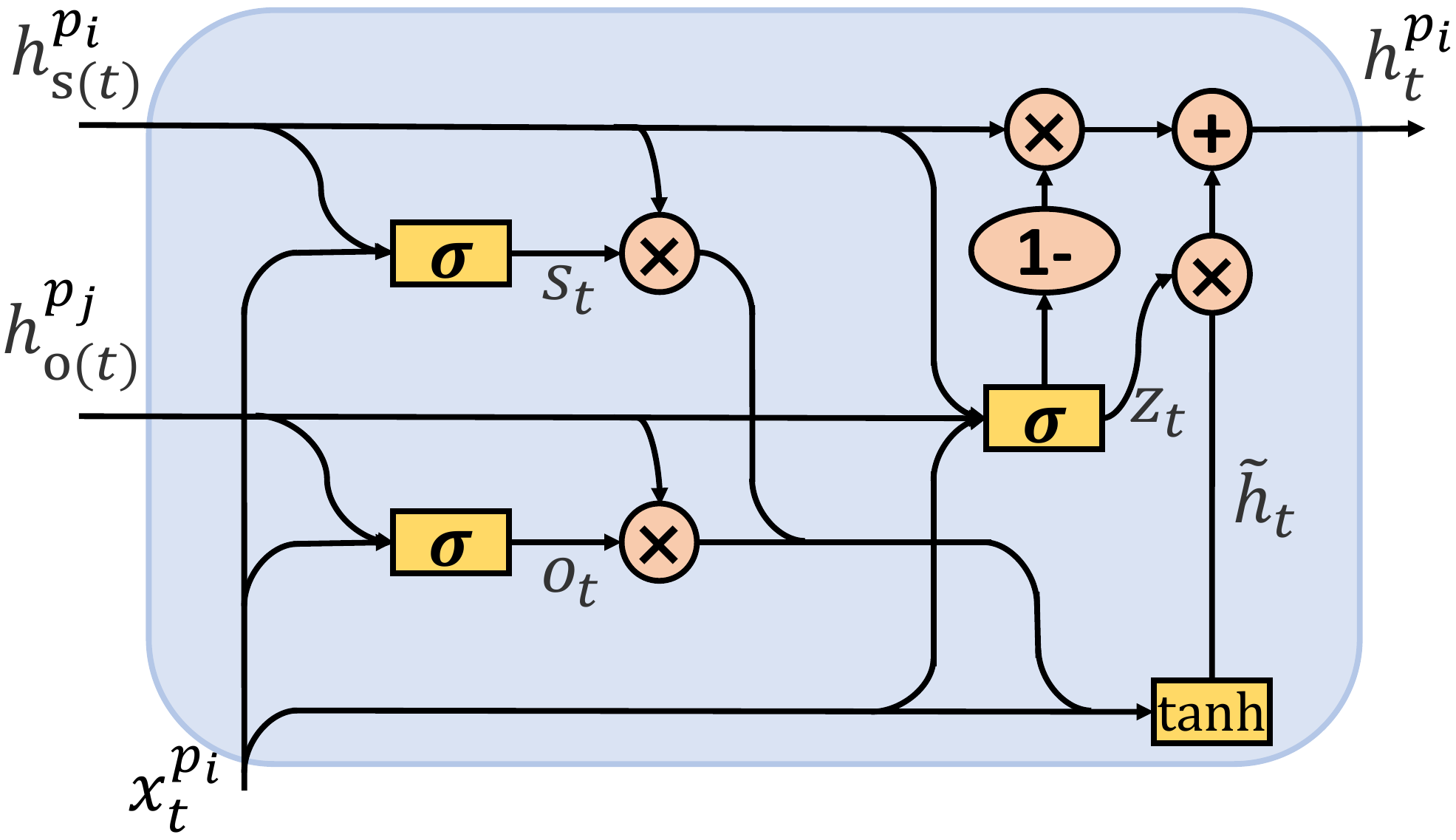}
		\label{fig:diagrucell}
	}
	\subfloat[]{\includegraphics[height=1.2in]{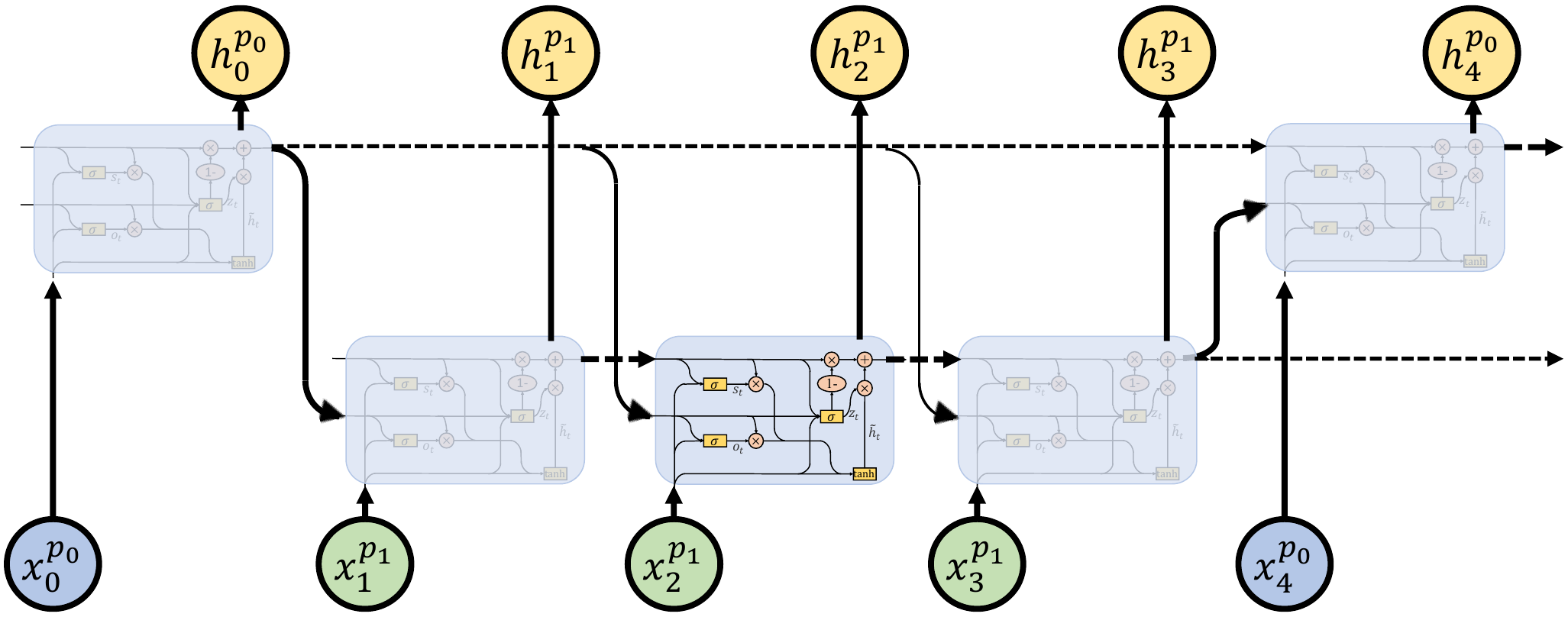}	
		\label{fig:diagru_example}
	}	
	\caption{Illustration of DiaGRU. (a) A single DiaGRU cell. Here, $x^{p_i}_{t}$, $h^{p_i}_{t}$, $h^{p_i}_{\bm{s}(t)}$, and $h^{p_j}_{\bm{o}(t)}$ are the vector representation of the utterance $u_t^{p_i}$, the hidden state of the current moment, the emotional hidden state of the speaker $p_i$, and the hidden state of the corresponding interlocutor $p_j$, respectively. (b) An example of DiaGRU. The dotted and solid lines represent self- and other-dependency, respectively, and the thickness indicates the strength of dependency adding decay factor.}
	\label{fig:diagrus}
\end{figure}
Emotional inertia and contagion in a conversation are often susceptible to temporal sequence. For example, the current utterance relies more on the near contexts than on the long-distance contexts. Although IMMHA captures all historical information about the participants, it fails to take into account the temporal sequential information in the conversation. In this subsection, we design a new Dialogue-based Gated Recurrent Unit (DiaGRU) network with reference to the structure of GRU~\cite{chung2014empirical} to aggregate the intra- and inter-speaker dependencies in the conversation. The architecture of a single DiaGRU cell is shown in Figure~\ref{fig:diagrucell}.

DiaGRU sets the self- and other-dependent reset gates. Firstly, a single DiaGRU cell calculates the corresponding forgetting degrees (i.e., reset gates) $s_t$ and $o_t$ based on the hidden state $h^{p_i}_{\bm{s}(t)}$ of self-context and the hidden state $h^{p_j}_{\bm{o}(t)}$ of others-context, respectively. Here, $s_t$ approaches $1$ indicates strong emotional inertia, while $o_t$ approaches $1$ indicates strong emotional contagion. Then, the candidate hidden state $\tilde{h}_t$ and update gate $z_t$ are generated by the joint calculation of $s_t$ and $o_t$. Finally, the hidden state $h_{t}^{p_i}$ is obtained by fusing candidate hidden states $\tilde{h}_t$ and $h^{p_i}_{\bm{s}(t)}$ based on update gate $z_t$. A single DiaGRU cell can be formalized as follows:
\begin{equation}
	\begin{aligned}
		&s_t 		= \sigma(W_{s}(x^{p_i}_{t}\oplus h^{p_i}_{\bm{s}(t)}) + b_{s}), \\ 
		&o_t 		= \sigma(W_{r}(x^{p_i}_{t}\oplus h^{p_j}_{\bm{o}(t)}) + b_{r}), \\ 
		&z_t 		= \sigma(W_{z}(x^{p_i}_{t}\oplus h^{p_i}_{\bm{s}(t)}\oplus h^{p_j}_{\bm{o}(t)}) + b_{z}), \\
		&\tilde{h}_t = \mathbf{tanh}(W_{h}(x^{p_i}_{t}\oplus (s_t \odot h^{p_i}_{\bm{s}(t)})\oplus (o_t\odot h^{p_j}_{\bm{o}(t)})) + b_h), \\ 
		&h^{p_i}_t 		= (1 - z_t)\odot h^{p_i}_{\bm{s}(t)} + z_t\odot\tilde{h}_t,
	\end{aligned}
	\label{eq:gru}
\end{equation}
where $\sigma(\cdot)$ is the sigmoid function, $\oplus$ denotes the concatenation operation; $s_t$ and $o_t$ are the weights of self- and other-dependent reset gates, respectively; $z_t$ is the weight of update gate; $W_s$, $W_r$, $W_z$, $W_h$, $b_r$, $b_s$, $b_z$, and $b_h$ are the learnable parameters.

In addition, as shown in Figure~\ref{fig:diagru_example}, a speaker-specific utterance block of the participant in a conversation may contain multiple utterances. If $h_{\bm{s}(t)}^{p_i}$ and $h_{\bm{o}(t)}^{p_j}$ are equally exerted on each DiaGRU cell, the others-dependence will dominate the hidden state $h_t^{p_i}$ at the current moment. Inspired by ECM~\cite{zhou2018emotional}, we apply two exponential decay factors, i.e., $\beta_{s,t}$ and $\beta_{o,t}$, for self- and other-dependency according to the time interval between two utterances. Our aim is that the longer the interval, the smaller the decay factor. The formula is as follows:
\begin{equation}
	\begin{aligned}
		&\beta_{s,t}   = \frac{1}{1+\exp\left[{(t-\bm{s}(t)-\mu_s)}/{\gamma_s}\right]},\\ 
		&\beta_{o,t}   = \frac{1}{1+\exp\left[{(t-\bm{o}(t)-\mu_o)}/{\gamma_o}\right]},\\ 		
		&h_t^{p_i} = \mathbf{DiaGRU}\left(x_t^{p_i}, (\beta_{s,t})h_{\bm{s}(t)}^{p_i}, (\beta_{o,t})h_{\bm{o}(t)}^{p_j}\right),
	\end{aligned}
\end{equation}
where $\mu_s$ and $\mu_o$ are the position hyperparameters, $\gamma_s$ and $\gamma_o$ are the shape hyperparameters, and these hyperparameters are utilized to control the speed of emotional decay; $\bm{s}(t)$ denotes the moment of the previous utterance that belongs to the same speaker as the utterance at moment $t$ and is the nearest to that utterance, and $\bm{o}(t)$ denotes the moment of the previous utterance that is spoken by the interlocutor and is the nearest to the utterance at moment $t$; $\mathbf{DiaGRU}(\cdot)$ is a simplified function of Equation~\ref{eq:gru}.

\subsection{Skip-chain conditional random field}
At the feature-extraction level, the global contextual features extracted by IMMHA and the local contextual features extracted by DiaGRU are combined to obtain identity-enhanced utterance features. Considering that there are significant dependencies between emotions in the conversation, we explicitly model the emotion interactions at the classification level leveraging CRF to capture emotional flows from different participants as well as to obtain the optimal sequence of emotions.

Because of the complexity of computing the normalized factor in the graphical model, only first-order dependency is commonly exploited in existing ERC methods~\cite{wang2020contextualized, song2022emotionflow}. In other words, only the dependency between direct neighboring utterances is taken into account. This simple processing makes it challenging to capture the complex interaction information of distinct speakers in the dialogue, while not clearly expressing the meaning represented by the non-normalized transition probability matrix in CRF. Therefore, by introducing higher-order dependency into the CRF, we elaborate a novel strategy named Skip-chain Conditional Random Field (SkipCRF) to model the emotion interactions in the dialogue and enhance the performance of emotion classification.

According to the relevant settings of the CRF, the value $\mathrm{x}=(x_1,x_2,...,x_\mathcal{T})$ is defined as the utterance feature extracted by encoders at the feature-extraction level, and $\mathrm{y}=(y_1,y_2,\cdots,y_\mathcal{T})$ is defined as the corresponding emotion, where $y_t \in E$. Referring to Equation~\ref{eq:linercrf}, $\bm{g}_i(\cdot)$ represents the contextual local feature function, and $\bm{f}_l(\cdot)$ is the nodal feature function. Subject to Markov property~\cite{lafferty2001,MAL-013}, the contextual local feature function $\bm{g}_i(\cdot)$ of the linear-chain CRF adopt only the neighboring information with unseen identity, resulting in its inability to distinguish between the impacts brought by the speaker and the interlocutor. In contrast, SkipCRF, subdivides contextual local feature function $\bm{g}_i(\cdot)$ into the self-dependent feature function $\bm{h}_i(y_{\bm{s}(t)},y_{t})$ and the others-dependent feature function $\bm{g}_j(y_{\bm{o}(t)},y_t)$ by introducing the speaker identity. The conditional probability $\mathbf{P}{(\mathrm{y}|\mathrm{x})}$ is defined as follows: 
\begin{equation}
	\begin{aligned}
		\mathbf{P}{(\mathrm{y}|\mathrm{x})}  &= \frac{1}{\mathbf{Z}(\mathrm{x})}\exp\big[\sum_{i,j,t}\left(\lambda_i \bm{h}_i(y_{\bm{s}(t)}, y_{t}, \mathrm{x}, t)  +  \eta_{j} \bm{g}_{j}(y_{\bm{o}(t)}, y_t,\mathrm{x},t)\right) +\sum_{l,t}\mu_l \bm{f}_l(y_t,\mathrm{x},t) \big] \\
		&= \frac{1}{\mathbf{Z}(\mathrm{x})}\exp\big[\sum_{n,t}\omega_n \bm{F}_n(y_{\bm{s}(t)},y_{\bm{o}(t)}, y_{t},\mathrm{x})\big ], 
	\end{aligned}
\end{equation}
where $\mathbf{Z}(\mathrm{x})$ is the normalization factor of all state sequences; $\omega_n$ consists of $\lambda_i$, $\eta_j$, and $\mu_l$, which represents the learnable weight of the feature function.

It is noted that CRF belongs to the probabilistic graphical model, and the difficulty lies in the calculation of the normalization factor with exponential complexity. Therefore, we employ forward-backward algorithm~\cite{binder1997space} to recursively calculate $\mathbf{P}{(\mathrm{y}|\mathrm{x})}$, making it linearly complex. Distinct from the linear-chain CRFs that have been applied in existing efforts~\cite{wang2020contextualized,song2022emotionflow,liang2021s+}, the designed SkipCRF distinguishes the identity information of the interlocutor. 
We define $\alpha_t(y_{\bm{o}(t)}^{p_j},y_t^{p_i}|\mathrm{x})$ as the non-normalized probability of partial emotion sequences before the moment $t$ when the speaker's emotion is $y_t^{p_i}$ and the interlocutor's emotion is $y_{\bm{o}(t)}^{p_j}$. Assuming the total number of possible emotion labels is $\mathcal{K}$, we define $\mathrm{A}_t(\mathrm{x})$ as a forward matrix/tensor consisting of $\mathcal{K}\times \mathcal{K}$ values:  %when the speaker's emotion is $y_t^{p_i}$ and the interlocutor's emotion is $y_{\bm{o}(t)}^{p_j}$ 
\begin{equation}
	\mathrm{A}_t(\mathrm{x})=
	\begin{bmatrix}
		\alpha_t(y_{\bm{o}(t)}^{p_j}=1, y_t^{p_i}=1|\mathrm{x}), &\cdots, &\alpha_t(y_{\bm{o}(t)}^{p_j}=1, y_t^{p_i}=\mathcal{K}|\mathrm{x})\\
		\vdots & \ddots &  \vdots\\
			\alpha_t(y_{\bm{o}(t)}^{p_j}=\mathcal{K}, y_t^{p_i}=1|\mathrm{x}), &\cdots, &\alpha_t(y_{\bm{o}(t)}^{p_j}=\mathcal{K}, y_t^{p_i}=\mathcal{K}|\mathrm{x})\\
	\end{bmatrix}.
	\label{eq:forwardmatrix}
\end{equation}
Here, the value of the $m$-th row and $n$-th column in $\mathrm{A}_t(\mathrm{x})$ denotes the non-normalized probability when the interlocutor's $m$-th emotion and the speaker's $n$-th emotion. Following the setting of the forward-backward algorithm, the non-normalized transition probability $\mathrm{m}_{t}$ is defined as:
\begin{equation}
	\mathrm{m}_{t}(y_{\bm{s}(t)},y_{\bm{o}(t)},y_{t},\mathrm{x}) = \exp\big[\sum_{k,t}\omega_k \bm{F}_k(y_{\bm{s}(t)},y_{\bm{o}(t)},y_{t},\mathrm{x})\big].
\end{equation}
Then, the $\mathcal{K}\times \mathcal{K}\times \mathcal{K}$-dimensional non-normalized transition probability tensor $\mathrm{M}_{t}$ is constructed through transition probability $\mathrm{m}_{t}$. Therefore, the recursive equations for the forward probability matrix $\mathrm{A}_{t}$ and backward probability matrix $\mathrm{B}_t$ are as follows:
\begin{equation}
\begin{aligned}
	&\mathrm{A}_{t}^\top(\mathrm{x}) = \mathrm{A}_{t-1}^\top(\mathrm{x})\mathrm{M}_{t}(\mathrm{x}),\\ 
	&\mathrm{B}_{t}(\mathrm{x}) = \mathrm{M}_{t+1}(\mathrm{x})\mathrm{B}_{t+1}(\mathrm{x}).\ 
\end{aligned}
\end{equation}
So the expression of the normalization factor $\mathbf{Z}(\mathrm{x})$ is as follows: 
\begin{equation}
	\mathbf{Z}(\mathrm{x}) = \boldsymbol{1}^\top \mathrm{A}_{\mathcal{T}}(\mathrm{x})   \boldsymbol{1} = \boldsymbol{1}^\top   \mathrm{B}_{1}(\mathrm{x})  \boldsymbol{1},
\end{equation}
where $\boldsymbol{1}$ denotes the $\mathcal{K}$-dimensional vector whose elements are all $1$.

In the training process, given the utterance feature $\mathrm{x}=(x_1,x_2,\cdots,x_\mathcal{T})$ and ground-truth label sequence ${\mathrm{y}}^*=(y_1^*,y_2^*,\cdots,y_\mathcal{T}^*)$, the objective is to maximize $\mathbf{P}{(\mathrm{y}^*|\mathrm{x})}$ of the ground-truth label sequence. It is converted into the following minimization objective by the negative logarithmic function: 
\begin{equation}
	\begin{aligned}
		\mathcal{L}(\Theta) =  -\mathbf{log}(\mathbf{P}(\mathrm{y}^*|\mathrm{x})),
	\end{aligned}
\end{equation}
where $\Theta$ consists of the classification parameter $\Theta_{cls}$ and the feature-extraction parameter $\Theta_{ext}$, and these two types of parameters can be updated by the back propagation algorithm.

After completing conditional probability modeling, the decoding problem of SkipCRF requires to be solved. Given the conditional probability $\mathbf{P}{(\mathrm{y}|\mathrm{x})}$ and the input feature $\mathrm{x}$, the emotion sequence $\mathrm{y}$ when $\mathbf{P}{(\mathrm{y}|\mathrm{x})}$ takes the maximum is obtained. We adopt the Viterbi algorithm~\cite{Forney1973Theviterbi} for solving the problem. Similar to the non-normalized probability $\alpha_t(y_{\bm{o}(t)}^{p_j},y_t^{p_i}|\mathrm{x})$, we define a local state $\delta_{t}(y_{\bm{o}(t)}^{p_j},y_{t}^{p_i})$, which represents the maximum non-normalized probability corresponding to all possible values of $y_{\bm{o}(t)}^{p_j}$ and $y_t^{p_i}$. Here, the normalization factor does not affect the comparison of maximum values, so only the non-normalization probability is needed. The recursive equation of local state can be formalized as follows: 
\begin{equation}
	\delta_{t}(\hat{k}_1,\hat{k}_2) = \mathop{\mathbf{max}}_{1\le k_1,k_2\le \mathcal{K}}\{\delta_{t-1}(k_1, k_2)
	+\sum_{n}\omega_n\bm{F}_n(y_{\bm{s}(t)}, y_{\bm{o}(t)}=\hat{k}_1,y_{t}=\hat{k}_2,\mathrm{x})\},
\end{equation}
where $k_1$, $k_2$, $\hat{k}_1$, and $\hat{k}_2$ denote emotions corresponding to utterances. In addition, when $\delta_{t}(y_{\bm{o}(t)}^{p_j},y_t^{p_i})$ reaches the maximum, the local state $\Psi_{t}({y}_{\bm{o}(t)}^{p_j},{y}_t^{p_i})$ is used to record emotion values of $y_{\bm{o}(t)}^{p_j}$ and $y_t^{p_i}$, which is applied to trace the optimal solution: 
\begin{equation}
	\Psi_{t}(\hat{k}_1,\hat{k}_2) =\mathop{\mathbf{argmax}}_{1\le k_1,k_2\le \mathcal{K}}\{\delta_{t-1}(k_1, k_2)
	+\sum_{n}\omega_n\bm{F}_n(y_{\bm{s}(t)} , y_{\bm{o}(t)}=\hat{k}_1 ,y_{t}=\hat{k}_2,\mathrm{x})\}.
\end{equation}

\begin{figure}[htbp]
	\centering
	\captionsetup[subfigure]{}
	\subfloat[]{\includegraphics[height=1.5in]{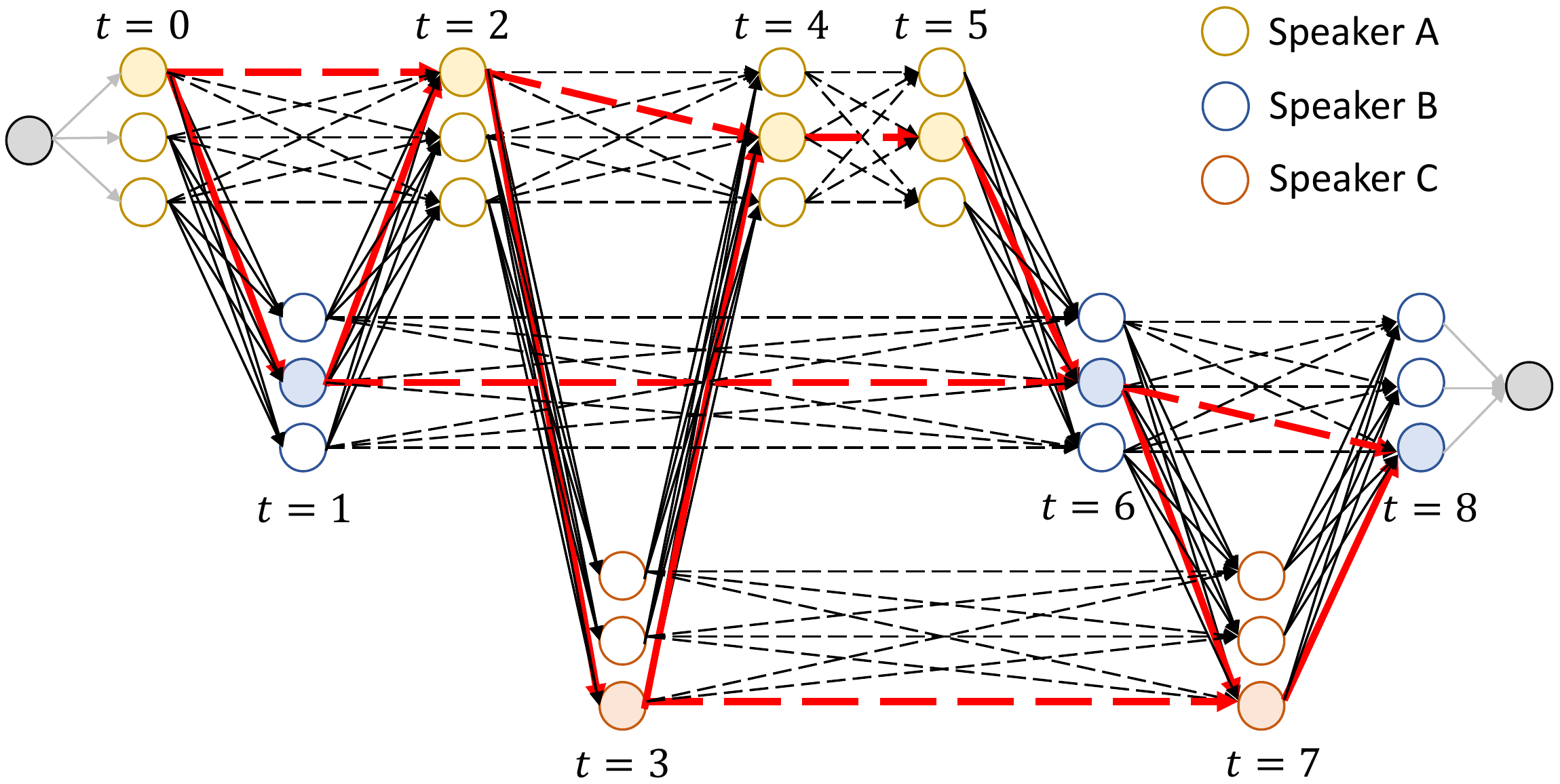}
		\label{fig:before_eliminate}
	}
	\subfloat[]{\includegraphics[height=1.5in]{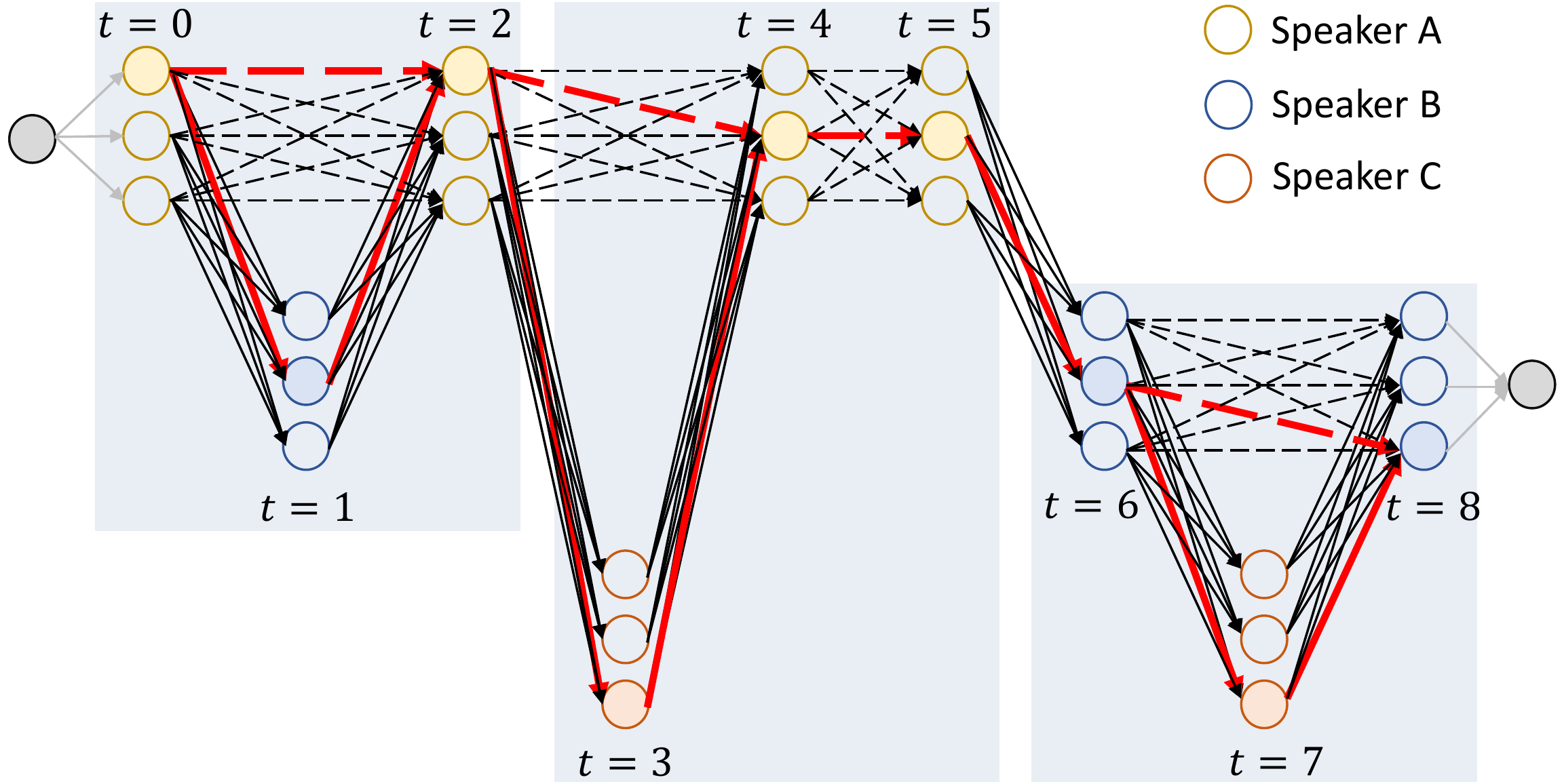}	
		\label{fig:after_eliminate}
	}	
	\caption{Example of CRF modeling for a three-person conversation scenario (a) before and (b) after eliminating skip-chain connections. The nodes with different colors represent different participants' node features. The dotted and solid lines represent the weights of self- and other-dependent features, respectively. The filled nodes represent the ground-truth labels, and the red line connects the optimal emotion sequence and represents the optimal path.}
	\label{fig:skipchain_eliminate}
\end{figure}
It is noted that the above analysis is discussed under the premise of dyadic conversation scenario. As can be noticed through Equation~\ref{eq:forwardmatrix}, the order of the forward tensor is equal to the number of participants, so the computational complexity increases exponentially with the number of participants, which is not conducive to multi-person conversation scenario. Therefore, we do some simplifications, i.e., eliminating those skip-chain connections that span the dyadic conversation in order to make it a multi-segment dyadic conversation. Note that this is still a full undirected graphical model. The specific example presented in Figure~\ref{fig:skipchain_eliminate} provides a visual illustration of the discrepancy before and after elimination. It can be seen from Figure~\ref{fig:after_eliminate} that the connection between moment $1$ and moment $6$, and the connection between moment $3$ and moment $7$ have been removed. The shaded area shows that the multi-party conversation is converted into a three-segment dyadic conversation after eliminating these connections. The implication is that only the interaction between two participants needs to be explicitly considered in the multi-party conversation, while the influence of other participants' utterances is implicitly encoded.

\section{Experimental settings}\label{sec:experimental_settings}
\subsection{Evaluation metrics and datasets}
\begin{table}[htbp]
    \centering
    \renewcommand{\arraystretch}{1.0}
    \setlength{\tabcolsep}{5pt}
    \caption{Statistics for these four emotion datasets. \#Conversation and \#Utterance denote the number of conversations and utterances, respectively.}
    \begin{tabular}{c|c|cccc}
	\toprule
    \multicolumn{2}{c|}{Datasets} &IEMOCAP &DailyDialog &MELD &EmoryNLP\\
    \hline
    \multirow{3}{*}{\#Conversation} &Train &108  &11,118 &1,039 &659\\
     &Validation &12 &1,000  &114 &89\\
     &Test &31 &1,000 &280 &79\\
    \hline
    \multirow{3}{*}{\#Utterance} &Train &5,163 &87,170 &9,989 &7,551\\
     &Validation &647 &8,069 &1,109 &954\\
     &Test &1,623 &7,740 &2,610 &984\\
	\bottomrule
    \end{tabular}
    \label{tab:statistics}
\end{table}
With reference to COSMIC~\cite{Ghosal2020}, we evaluate EmotionIC on four benchmark ERC datasets including IEMOCAP~\cite{busso2008iemocap}, DailyDialog~\cite{li2017dailydialog}, MELD~\cite{poria2019meld}, and EmoryNLP~\cite{zahiri2017emotion}. The statistics of these datasets are reported in Table~\ref{tab:statistics}. For the IEMOCAP dataset, we choose the weighted F1 and accuracy to validate the proposed model; for the DailyDialog dataset, the macro F1 and micro F1 excluding the majority class (\texttt{Neutral}) are utilized to evaluate our model; for the MELD and EmoryNLP datasets, we select the weighted F1 and micro F1. In addition, all data segmentation and pre-processing is consistent with COSMIC~\cite{Ghosal2020}.

\subsection{Implementation details}
\begin{table}[htbp]
    \centering
    \renewcommand{\arraystretch}{1.0}
    \setlength{\tabcolsep}{4.0pt}
    \caption{Partial hyperparameter settings for distinct datasets. LR-ext and LR-cls denote the learning rates for the feature-extraction and classification levels, respectively. BS and DR are the batch size and dropout rate, respectively. D-IMMHA and D-DiaGRU indicate the network depths for IMMHA and DiaGRU modules, respectively.}
    \begin{tabular}{c|ccccccc}
	\toprule
    Datasets &LR-ext &LR-cls &L2 &BS &DR &D-IMMHA &D-DiaGRU \\
	\hline
    IEMOCAP &2e-05 &7e-03 &1e-04 &32 &0.3 &5 &3 \\
	DailyDialog &2e-05 &1e-03 &1e-04 &128 &0.3 &6 &3 \\
    MELD &5e-05 &9e-03 &1e-04 &128 &0.3  &3 &3 \\
    EmoryNLP &3e-05 &1e-04 &1e-04 &128 &0.4  &4 &2 \\
    \bottomrule
    \end{tabular}
    \label{tab:hyperparameter}
\end{table}
In this subsection, we mainly describe the implementation details of our proposed method. Since our approach considers only conversation-level modeling, the RoBERTa model fine-tuned by utterance classification task is used to extract utterance-level features with the dimensionality of 1024 as input in advance~\cite{Ghosal2020}. The maximum epoch is set to 100. $\mu_s$, $\mu_o$, $\gamma_s$ and $\gamma_o$ are set to 3, 0, 1, and 2, respectively. During training, the Adam optimizer~\cite{loshchilov2018decoupled} is adopted for optimization. All experiments are conducted on a server with an NVIDIA GeForce RTX 3090. The partial hyperparameter settings are shown in Table~\ref{tab:hyperparameter}.

\section{Results and analysis}\label{sec:results_analysis}
\subsection{Comparison with baseline methods}
\begin{table}[htbp]
\caption{Experimental results of EmotionIC and baseline models on four datasets. All baseline results are from the original paper and only textual modality is used.}
\label{tab:main} 
\begin{center}
\renewcommand{\arraystretch}{1.0}
\footnotesize
\setlength{\tabcolsep}{4.5pt}
\begin{tabular}{c|cc|cc|cc|cc}
\toprule
%			\hline
\multicolumn{1}{l|}{\multirow{3}{*}{Models}} &  \multicolumn{2}{c|}{IEMOCAP} &  \multicolumn{2}{c|}{DailyDialog} &  \multicolumn{2}{c|}{MELD}  &  \multicolumn{2}{c}{EmoryNLP} \\
\cline{2-9}
& \multicolumn{1}{c}{\multirow{2}{*}{\tabincell{c}{Weighted-F1}}} & \multicolumn{1}{c|}{\multirow{2}{*}{Accuracy}} & \multicolumn{1}{c}{\multirow{2}{*}{\tabincell{c}{Macro-F1}}} & \multicolumn{1}{c|}{\multirow{2}{*}{\tabincell{c}{Micro-F1}}} & \multicolumn{1}{c}{\multirow{2}{*}{\tabincell{c}{Weighted-F1}}} & \multicolumn{1}{c|}{\multirow{2}{*}{Micro-F1}}  & \multicolumn{1}{c}{\multirow{2}{*}{\tabincell{c}{Weighted-F1}}} & \multicolumn{1}{c}{\multirow{2}{*}{Micro-F1}} \\
& & & & & & & &\\
\hline
COSMIC 		& 65.28 & - 	& 51.05 & 58.48 & 65.21 & - 	& 38.11 & -\\
RGAT-ERC		&65.22 &- & - & 54.31 &60.91 &- &34.42 &- \\		
DialogXL	&65.94 &- & - & 54.93 &62.41 &- &34.73 &- \\
DAG-ERC 	& 68.03 & - 	& - 	& 59.33 & 63.65 & -  	& 39.02 & -\\

I-GCN       &66.28 &- &- &- &65.74 &- &- &-\\
LR-GCN       &68.30 &68.50 &- &- &65.60 &- &- &-\\
CauAIN      &67.61 &- &53.85 &58.21 &65.46 &- &- &-\\

GAR-Net 	& 67.41 & - 	& 45.81 	& 56.97 & 62.11 & -  	& - & -\\
CoG-BART 	& 66.18 & 66.71 	& - 	& 56.29 & 64.81 & 65.95  	& 39.04 & 42.58\\
EmoCaps-Text 	& 69.49 & - 	& - 	& - & 63.51 & -  	& - & -\\	
\hline
{EmotionIC} & \textbf{69.61} 	& \textbf{69.44} 	& \textbf{54.19} & \textbf{60.13} 	& \textbf{66.32}	& \textbf{67.59} & \textbf{40.25} 		& \textbf{44.31}\\								
\bottomrule
\end{tabular}
\end{center}
\end{table}
To evaluate the effectiveness of our EmotionIC, our chosen baselines include COSMIC~\cite{Ghosal2020}, RGAT-ERC~\cite{wang2020relational}, DialogXL~\cite{shen2021dialogxl}, DAG-ERC~\cite{shen-etal-2021-directed}, I-GCN~\cite{nie2022igcn}, LR-GCN~\cite{ren2022lrgcn}, CauAIN~\cite{zhao2022cauain}, GAR-Net~\cite{xu2022gar}, CoG-BART~\cite{li2022contrast}, and EmoCaps~\cite{li2022emocaps}. Table~\ref{tab:main} shows the performance comparison of the proposed EmotionIC with all baseline methods. The experimental results demonstrate that our EmotionIC outperforms all baseline methods and markedly exceeds in some indicators.
\begin{itemize}
	\item[(1)] IEMOCAP. The average length of conversation in the IEMOCAP dataset is far more than the other datasets, thus this dataset contains richer contextual information. Compared to recurrence-based methods in Table~\ref{tab:main}, EmotionIC achieves great improvements, demonstrating its ability to capture long-distance context dependencies. Coupled with the use of DiaGRU to fine-tune local intra- and inter-speaker dependencies, which enables our model to achieve optimal performance.
	\item[(2)] DailyDialog. As with the IEMOCAP dataset, the DailyDialog dataset contains dyadic conversations. However, the difference is that this dataset suffers from a severe class imbalance, where the \texttt{Neutral} accounts for $83\%$. Coupled with the much shorter conversation length, the performance improvement of our model on the DailyDialog dataset is insignificant.
	\item[(3)] MELD. The utterances in the MELD dataset are much shorter than those in the IEMOCAP dataset, which means that emotional modeling is highly context-dependent. In addition, the fact that there are often two or more speakers in a conversation and that the amount of statements uttered by each participant is relatively small makes it difficult to model based on emotional inertia and contagion. Our EmotionIC reaches the best F1 score, even surpassing that of COSMIC which adds additional commonsense knowledge. We attribute the improvement to our contextual modeling at the classification level, i.e., SkipCRF.
	\item[(4)] EmoryNLP. Compared to baseline models, our EmotionIC achieves competitive performance on the EmoryNLP dataset. However, compared with on the MELD dataset, EmotionIC shows limited improvement on the EmoryNLP dataset for the same problem. The probable reason may be that EmoryNLP requires more commonsense knowledge.
\end{itemize}

\subsection{Analysis for confusion matrices}
\begin{figure}[htbp]
	\centering
	\subfloat[]{\includegraphics[height=2.0in]{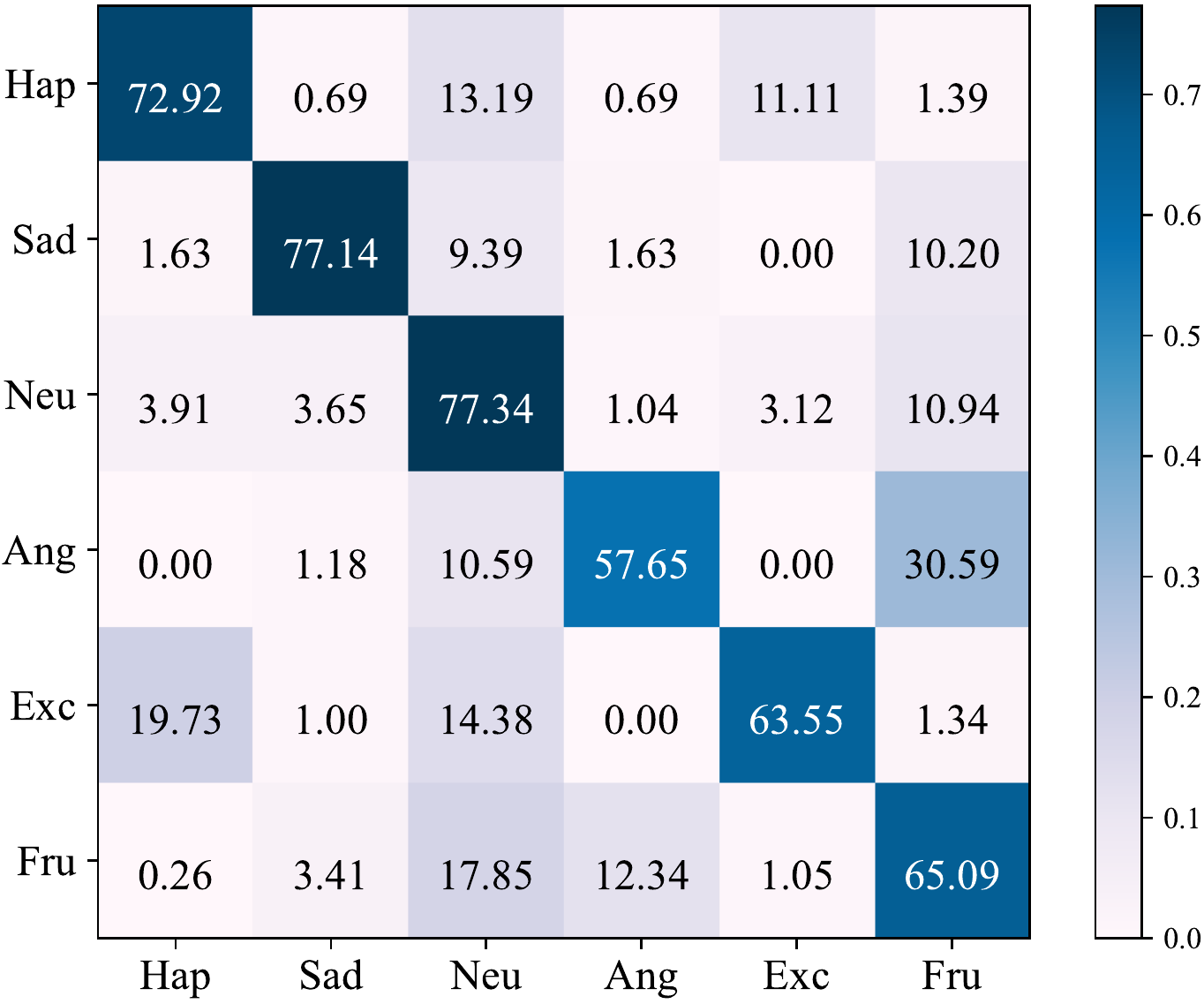}\label{fig_iemocap}
	}
	\hfil
	\subfloat[]{\includegraphics[height=2.0in]{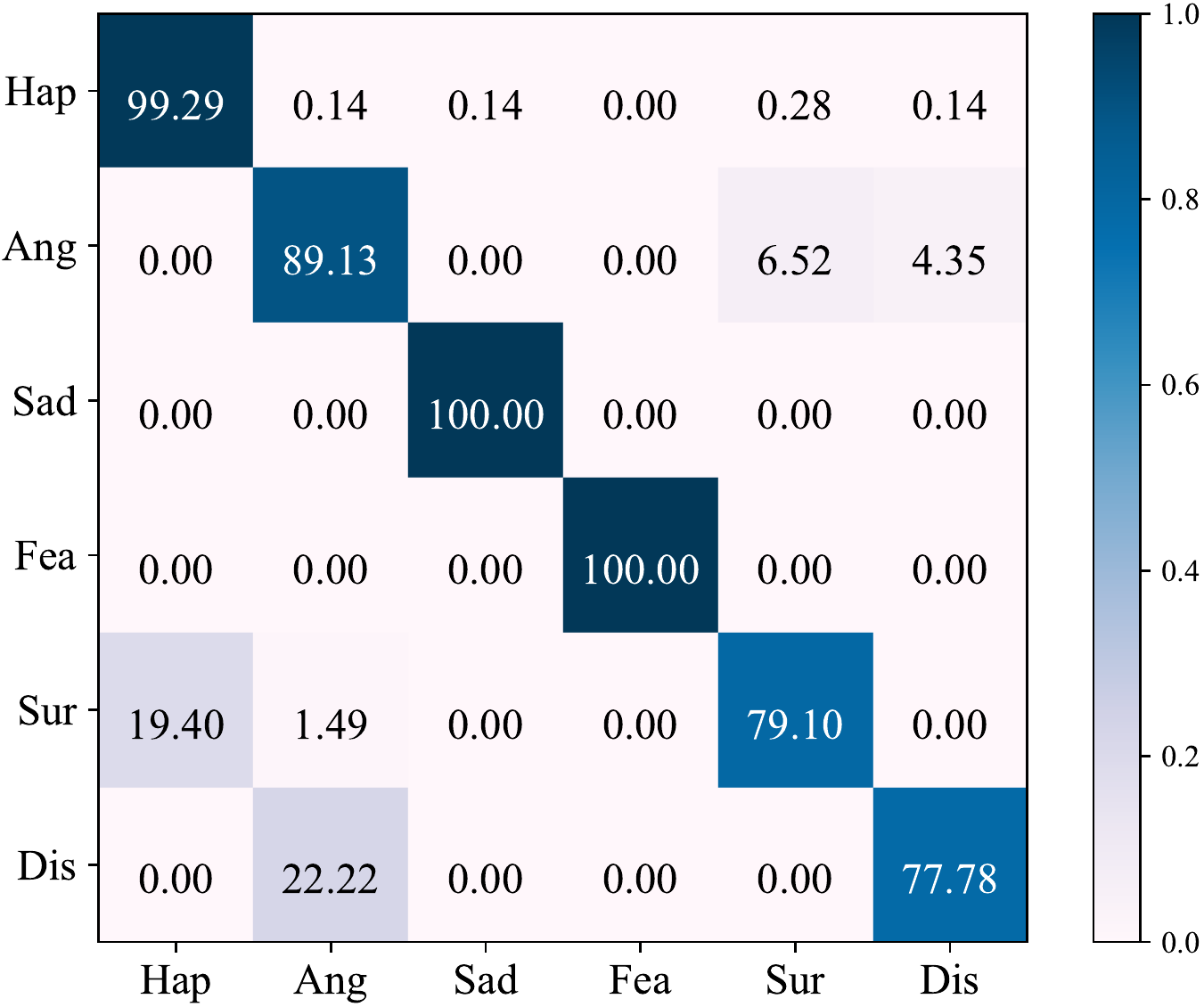}\label{fig_dailydialog}
	}
	% \hfil
	\\
	\subfloat[]{\includegraphics[height=2.0in]{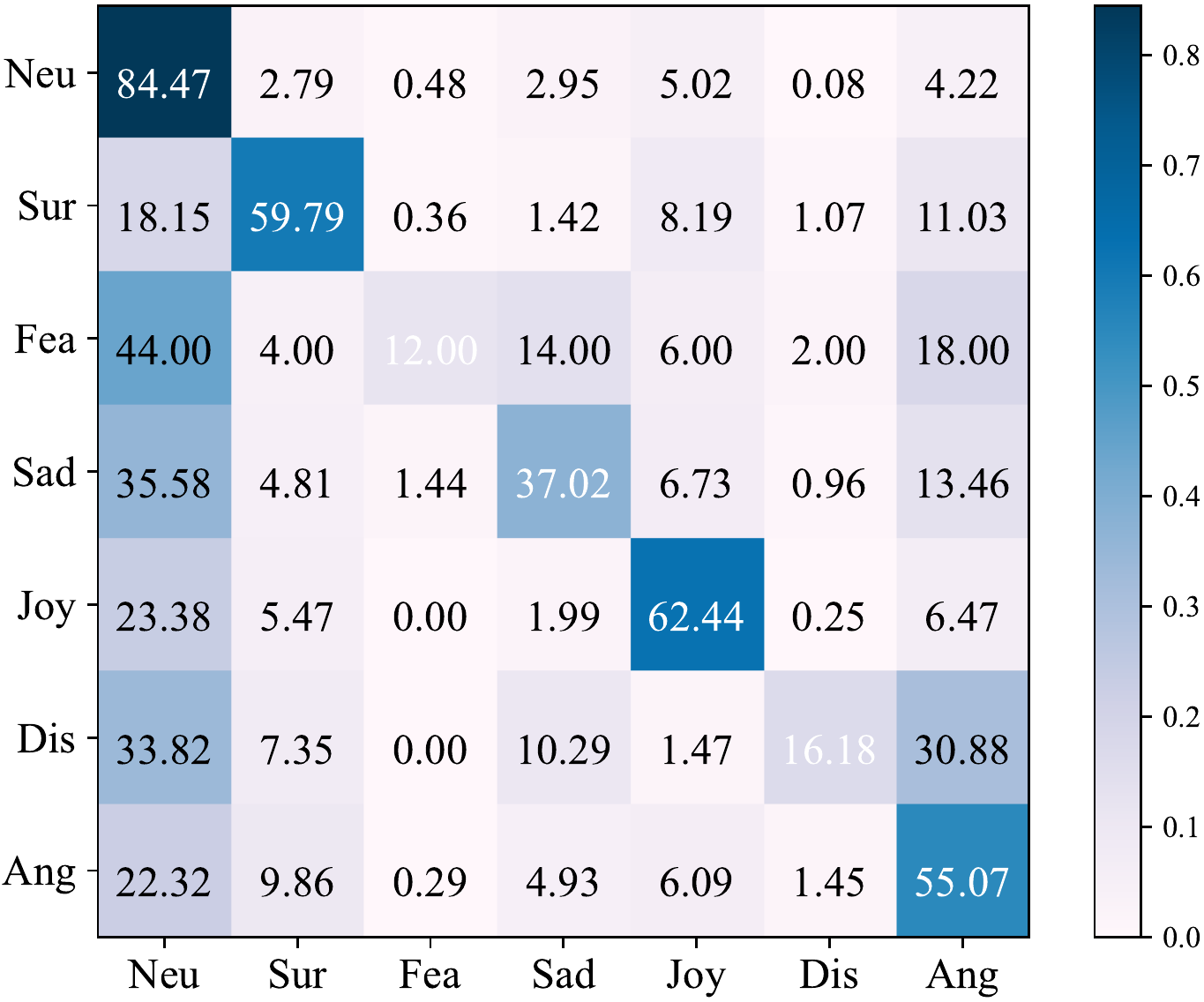}\label{fig_meld}
	}
	\hfil
	\subfloat[]{\includegraphics[height=2.0in]{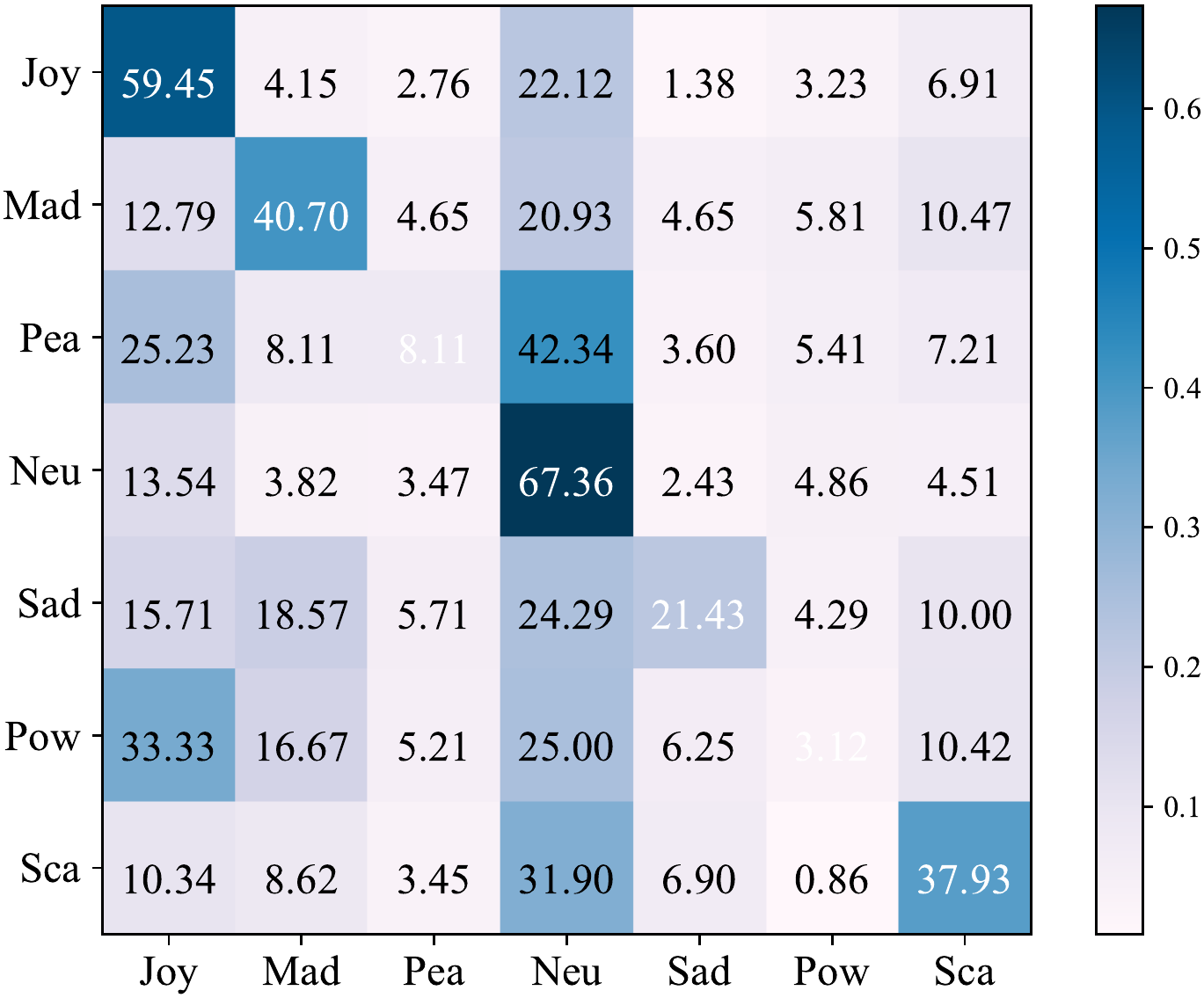}\label{fig_emorynlp}
	}	
	\caption{Confusion matrices of the testing set on the (a) IEMOCAP, (b) DailyDialog, (c) MELD, and (d) EmoryNLP datasets. The horizontal and vertical coordinate indicate the ground-truth and predicted emotion, respectively. Note that the majority class \texttt{Neutral} in the DailyDialog dataset is removed.}
	\label{fig_confusion}
\end{figure}
To further analyze the performance of EmotionIC, we show the confusion matrices for the four benchmark datasets in Figure~\ref{fig_confusion}. Overall, the proposed EmotionIC performs well on the IEMOCAP and DailyDialog datasets. For the IEMOCAP dataset, our EmotionIC can accurately recognize the emotions of utterances in most scenarios, exhibiting superior performance. It suggests that our proposed model can effectively capture intra- and inter-speaker contextual dependencies based on emotional inertia and contagion. In order to exclude the effect of extreme majority class, we remove the records related to \texttt{Neutral} in the DailyDialog dataset. It can be seen that our model has high accuracy. Even though the class number of \texttt{Happy} accounts for a high proportion in the DailyDialog dataset, EmotionIC can still distinguish it easily from other emotions.

In comparison to the first two datasets, the performance of the model on the MELD and EmoryNLP datasets is somewhat unsatisfactory. We examine the datasets and find that the MELD and EmoryNLP datasets are dialogue segments extracted from the TV series \textit{Friends}. In these two datasets, the length of most conversations is short and the two neighboring conversations may not be consecutive. Therefore, it is difficult for EmotionIC to exploit the capability of contextual modeling. In future studies, we will explore the use of commonsense knowledge to solve this problem. In addition, our model suffers from class imbalance and similar emotion problems. On the MELD dataset, \texttt{Fear} and \texttt{Disgust} belong to the minority classes, so, like most ERC models, EmotionIC has difficulty identifying them correctly. On the IEMOCAP dataset, \texttt{Angry} is easily recognized as similar emotion \texttt{Frustrated} in some scenarios.

\subsection{Result for each emotion}
\begin{figure}[htbp]
\centering
\subfloat[]{\includegraphics[height=2.0in]{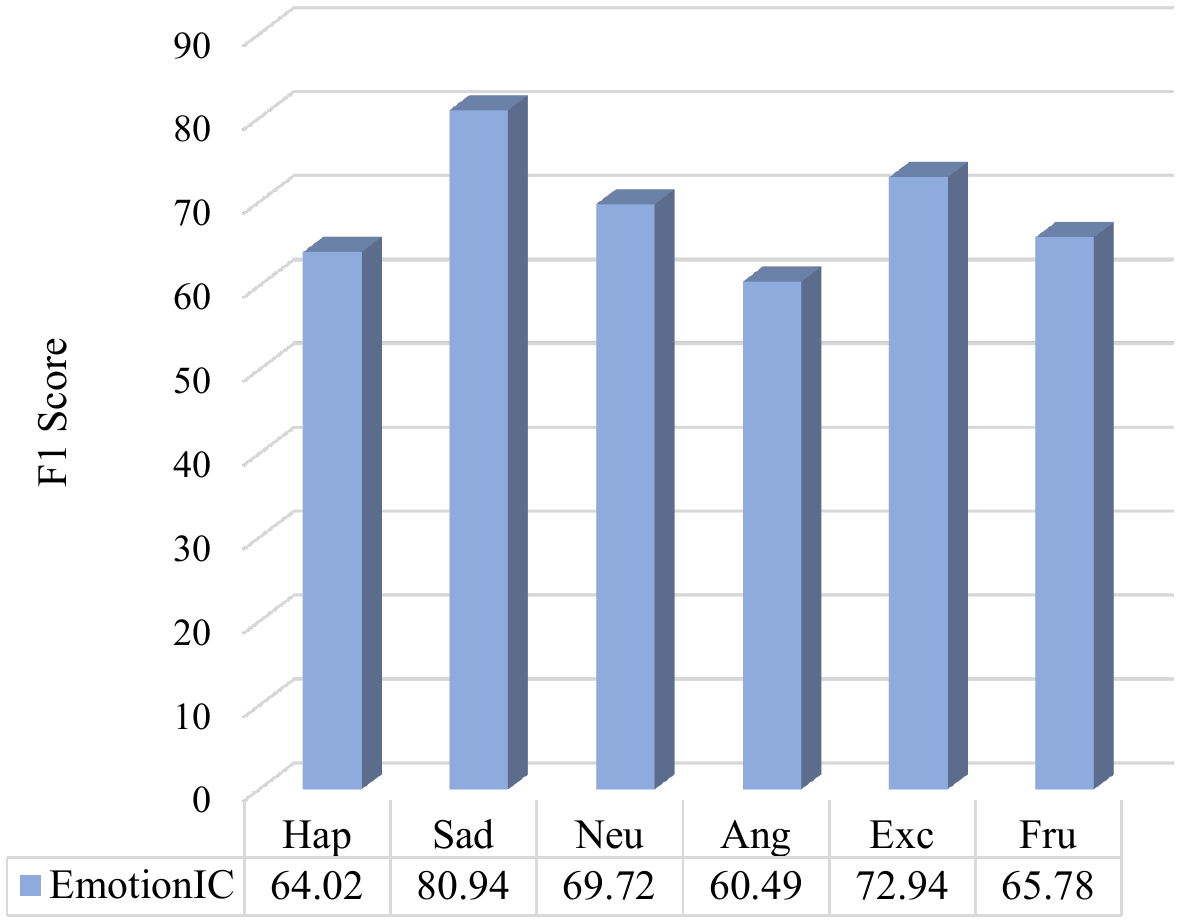}
	\label{each_iemocap}
}
\hfil
\subfloat[]{\includegraphics[height=2.0in]{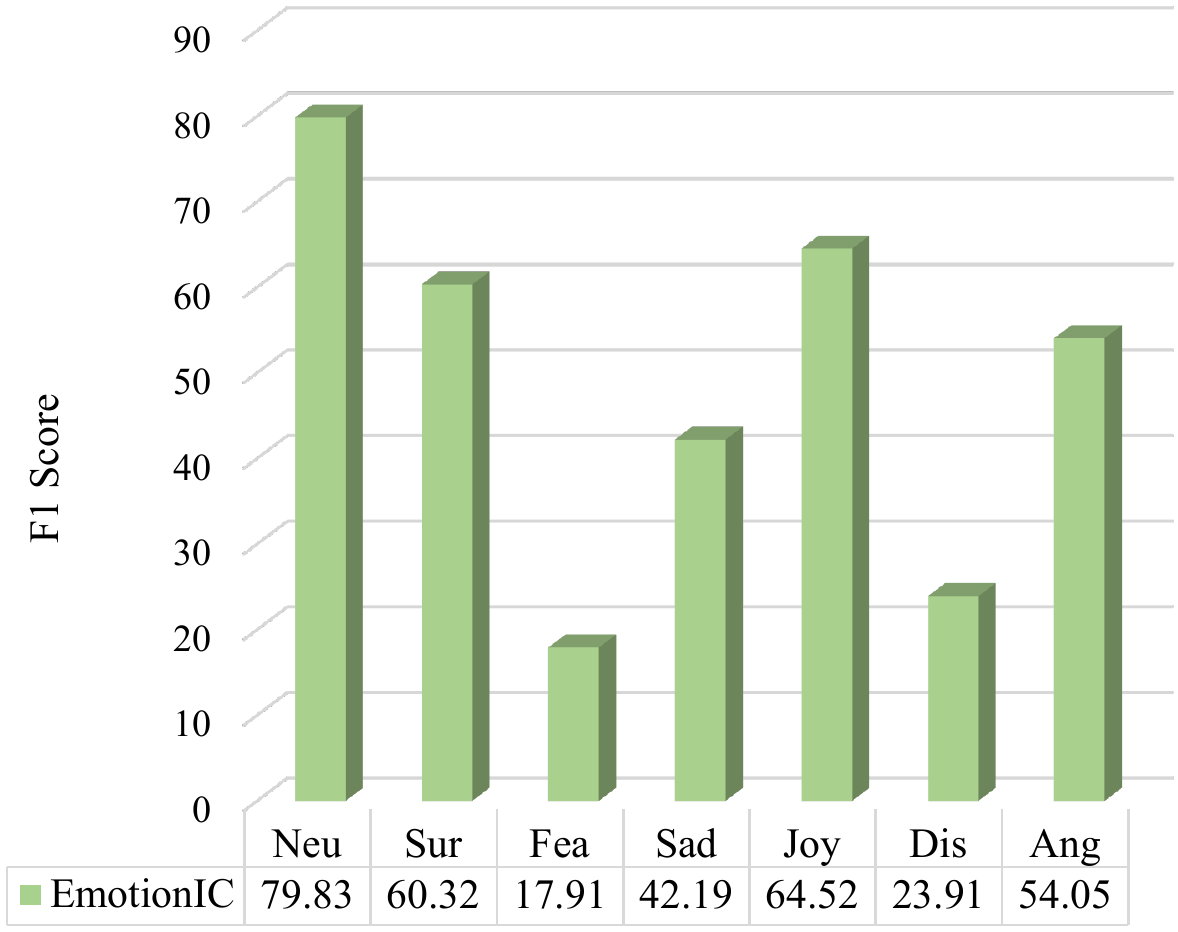}	
	\label{each_meld}
}
\caption{F1 score for each emotion on the (a) IEMOCAP and (b) MELD datasets. Note that Hap represents for the first three-letter abbreviation for \texttt{Happy}, and so on for other emotions.}
\label{fig:eachemotion}
\end{figure}
Figure~\ref{fig:eachemotion} shows the F1 scores of EmotionIC for each emotion on the IEMOCAP and MELD datasets. On the IEMOCAP dataset, the emotion that achieves the highest F1 score is \texttt{Sad}, which suggests that this class is more easily distinguished by EmotionIC relative to the other emotions. However, the F1 scores for \texttt{Angry} and \texttt{Happy} are lower in comparison to the other emotions, indicating that these two emotions are susceptible to being recognized as other emotions such as \texttt{Frustrated}. On the MELD dataset, \texttt{Neutral} achieves the highest F1 score by a significant margin, while \texttt{Fear} and \texttt{Disgust} obtain extremely low results. By examining the class distribution of MELD, we notice that the sample share (approximately 46.95\%) of \texttt{Neutral} in the dataset is far higher than that of \texttt{Fear} or \texttt{Disgust}. Therefore, we believe that the above phenomenon is due to the class imbalance problem. On the whole, the performance of EmotionIC for each class on the IEMOCAP dataset is more balanced due to the influence of the class imbalance in the MELD dataset.

\subsection{Ablation studies}
In order to analyze the impact of different modules in EmotionIC, we observe the performance after removing or replacing each module in this subsection. The experimental results are recorded in Tables~\ref{tab:modules} and~\ref{tab:modules_}. Overall, removing or replacing any of these modules leads to the performance degradation of EmotionIC, which demonstrates that our designed modules help to extract the contextual dependencies adequately.

\begin{table}[htbp]
	\centering	
	\caption{Impact of removing different modules. - indicates removal of the corresponding module.}
	\label{tab:modules}	
	\renewcommand{\arraystretch}{1.0}
	\setlength{\tabcolsep}{9pt}	
	\begin{tabular}{cc|c|c|c|c}
	\toprule
	%			\hline
	\multicolumn{2}{c|}{\multirow{2}{*}{Methods}} &\multicolumn{1}{c|}{IEMOCAP}& \multicolumn{1}{c|}{DailyDialog} & \multicolumn{1}{c|}{MELD} & \multicolumn{1}{c}{EmoryNLP}\\
	\cline{3-6}
	& &Weighted-F1 & Micro-F1 &Weighted-F1 &Weighted-F1\\
	\hline
	\multicolumn{2}{c|}{EmotionIC}	 			& \textbf{69.61} & \textbf{60.13} & \textbf{66.32} & \textbf{40.25} \\
	\hline 
	\multicolumn{2}{c|}{-IMMHA}				    & 66.61  ($\downarrow$3.00)  &58.35 ($\downarrow$1.78)& 65.77 ($\downarrow$0.55) &38.26 ($\downarrow$1.99)\\ 
	\multicolumn{2}{c|}{-DiaGRU}	          & 67.92  ($\downarrow$1.69)&59.78 ($\downarrow$0.35)& 65.95  ($\downarrow$0.37) &38.28 ($\downarrow$1.97)\\
	\hline
	\multicolumn{1}{c|}{-Mask Matrix $M_s$} &{\multirow{2}{*}{IMMHA}} & 67.88  ($\downarrow$1.73)  &59.44 ($\downarrow$0.69)& 65.92 ($\downarrow$0.40) &38.83 ($\downarrow$1.42)\\ 
	\multicolumn{1}{c|}{-Mask Matrix $M_o$} &  &  69.17 ($\downarrow$0.44)  &58.49 ($\downarrow$1.64)& 65.93 ($\downarrow$0.39) &39.12 ($\downarrow$1.13)\\
	\hline
	\multicolumn{1}{c|}{-Reset Gate $s_t$} &{\multirow{2}{*}{DiaGRU}}& 69.47 ($\downarrow$0.14) &59.99 ($\downarrow$0.14)& 65.52  ($\downarrow$0.80) &38.87 ($\downarrow$1.38)\\  	
	\multicolumn{1}{c|}{-Reset Gate $o_t$} &  &  68.92 ($\downarrow$0.69) &59.60 ($\downarrow$0.53)& 66.17  ($\downarrow$0.15) &39.08 ($\downarrow$1.17)\\
	\bottomrule
	\end{tabular}
	% }
\end{table}
Table~\ref{tab:modules} displays the experimental results after the removal of each module, from which the following findings can be drawn.
(1) On all datasets, removing IMMHA causes more performance degradation than removing DiaGRU. This indicates that capturing global context dependencies is more crucial than capturing local context dependencies. A major reason for this is that global context modeling takes into account long-distance information in the conversation.
(2) In comparison to the other datasets, removing IMMHA leads to the most performance degradation on the IEMOCAP dataset. This indicates that IEMOCAP relies more on global context modeling. This is due to the fact that IEMOCAP is a long-conversation dataset and relies more on the capability of IMMHA to capture long-distance information.
(3) Removing both the self- and other-dependent mask matrices of IMMHA results in decreased model performance. This suggests that explicitly extracting self- and other-dependency based on emotional inertia and contagion can effectively upgrade the performance of the model. The same conclusion can be drawn after removing the self- and other-dependent reset gates of DiaGRU.

\begin{table}[htbp]
	\centering	
	\caption{Impact of replacing different modules. - indicates removal of the corresponding module, + indicates replacement with another module.}
	\label{tab:modules_}	
	\renewcommand{\arraystretch}{1.0}
	\setlength{\tabcolsep}{9pt}	
	\begin{tabular}{c|c|c|c|c}
	\toprule
	%			\hline
	\multirow{2}{*}{Methods} &\multicolumn{1}{c|}{IEMOCAP}& \multicolumn{1}{c|}{DailyDialog} & \multicolumn{1}{c|}{MELD} & \multicolumn{1}{c}{EmoryNLP}\\
	\cline{2-5}
	&Weighted-F1 & Micro-F1 &Weighted-F1 &Weighted-F1\\
	\hline
	EmotionIC & \textbf{69.61} &\textbf{60.13} & \textbf{66.32} &\textbf{40.25} \\
	\hline 
	-IMMHA +MHA & 67.57  ($\downarrow$2.04)&57.85 ($\downarrow$2.28)& 65.50 ($\downarrow$0.82) &38.17  ($\downarrow$2.08)\\
	-DiaGRU +GRU & 66.31  ($\downarrow$3.30)&59.48 ($\downarrow$0.65)& 65.79 ($\downarrow$0.53) &39.02  ($\downarrow$1.23)\\
	-SkipCRF +Softmax &68.13  ($\downarrow$1.48)	&56.86 ($\downarrow$3.27)&63.99 ($\downarrow$2.33) &39.56 ($\downarrow$0.69)\\
	-SkipCRF +Linear-chain CRF & 69.17  ($\downarrow$0.44)&59.78 ($\downarrow$0.35)& 66.01 ($\downarrow$0.31) &39.18  ($\downarrow$1.07)\\	
	\bottomrule
	\end{tabular}
	% }
\end{table}
We record the F1 scores after replacing each module in Table~\ref{tab:modules_}, from which the following conclusions can be derived.
(1) Since MHA is unable to utilize the identity information of participants, it cannot sufficiently model global contexts.
(2) GRU fails to distinguish between intra- and inter-speaker dependency information, leading to low performance of the model.
(3) The direct use of softmax layer cannot explicitly mimic emotional propagations in the conversation and cannot effectively mine emotional flows.
(4) Linear-chain CRF focuses only on first-order dependencies at the classification level and cannot explicitly mine emotional flows between indirect neighbors. Meanwhile, Linear-chain CRF does not introduce participant identity to distinguish the influence of distinct speakers on the current utterance.

\subsection{Effectiveness of SkipCRF}
\begin{figure}[htbp]
	\centering
	\includegraphics[height=2.5in]{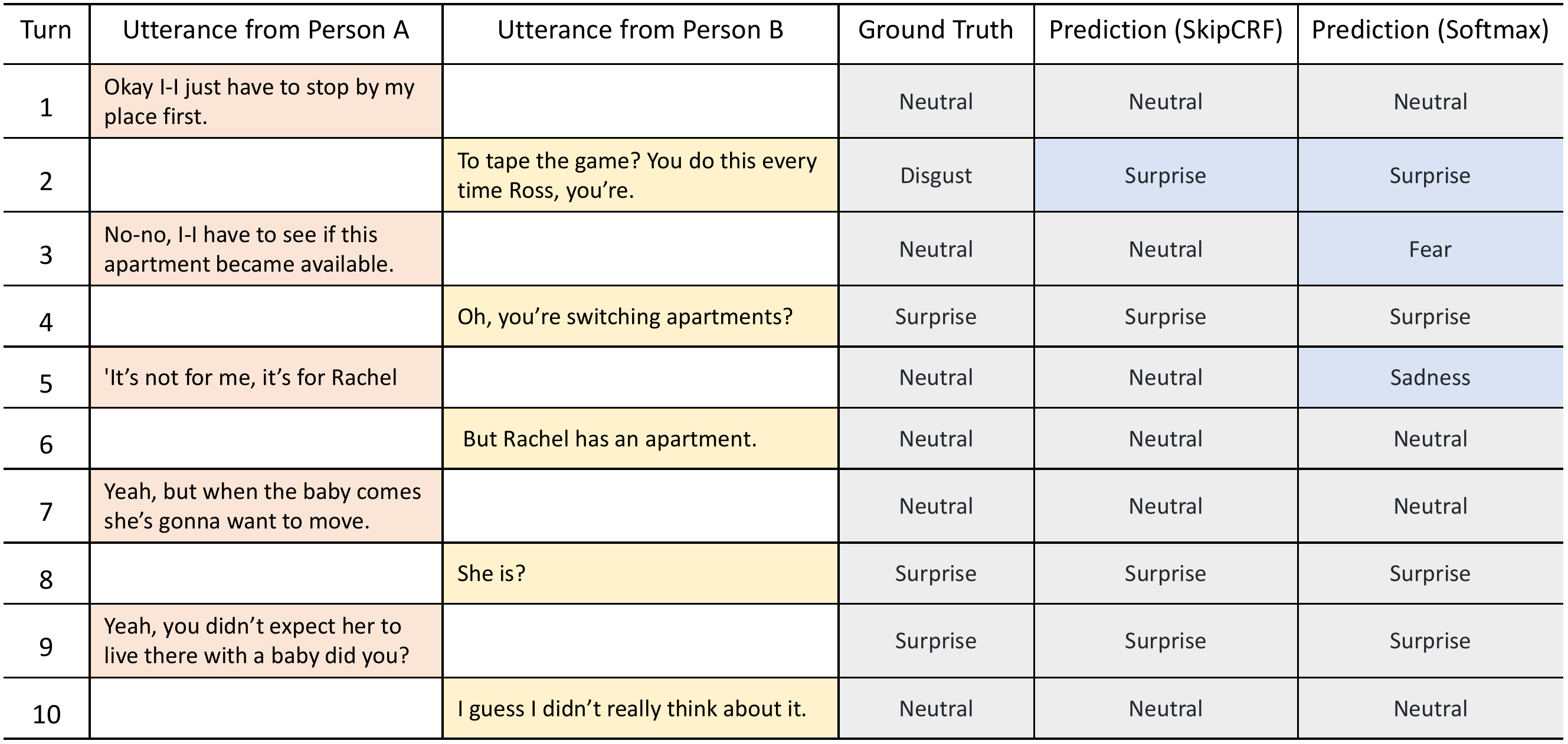}
	\caption{Case study on the MELD dataset. Prediction (SkipCRF) denotes the use of SkipCRF for emotion classification, and Prediction (Softmax) denotes the replacement of SkipCRF with the Softmax layer for emotion classification.}
	\label{fig:casestudy}
\end{figure}

To further prove that considering emotional flows in the conversation at the classification level can effectively improve the performance of the model, we conduct a case study with a conversation in the MELD dataset, as shown in Figure~\ref{fig:casestudy}. 
The dialogue presents emotional contagion between participants and their own emotional inertia. It can be observed that the first utterance (Turn 2) of Person B is incorrectly classified as \texttt{Surprise} due to the lack of reliable historical emotional information. From the misclassification of the $3$-th and $5$-th utterances, \texttt{Neutral} emotion is easily misclassified as negative one by the model employing Softmax layer. Our elaborate SkipCRF has significant advantages in modeling based on emotional inertia and contagion, proving the effectiveness of capturing emotional flows between different speakers at the classification level.

\subsection{Class imbalance in both the dialogue and dataset}
We conduct experiment on the DailyDialog dataset to investigate the impact of the class imbalance in the dialogue and that in the dataset together on the performance. Since the ERC model tends to be processed in terms of dialogues, the class imbalance problem in the dialogue affects its performance to some extent. For the emotion with a small sample size, if it is uniformly distributed in the dataset, i.e., it appears only a few times in each conversation. It is difficult for the model to obtain enough contextual information to express that emotion. On the contrary, if the emotion appears concentrated in several conversations, the model can effectively model based on emotional inertia and contagion.

\begin{figure}[htbp]
	\centering
	\includegraphics[height=2.0in]{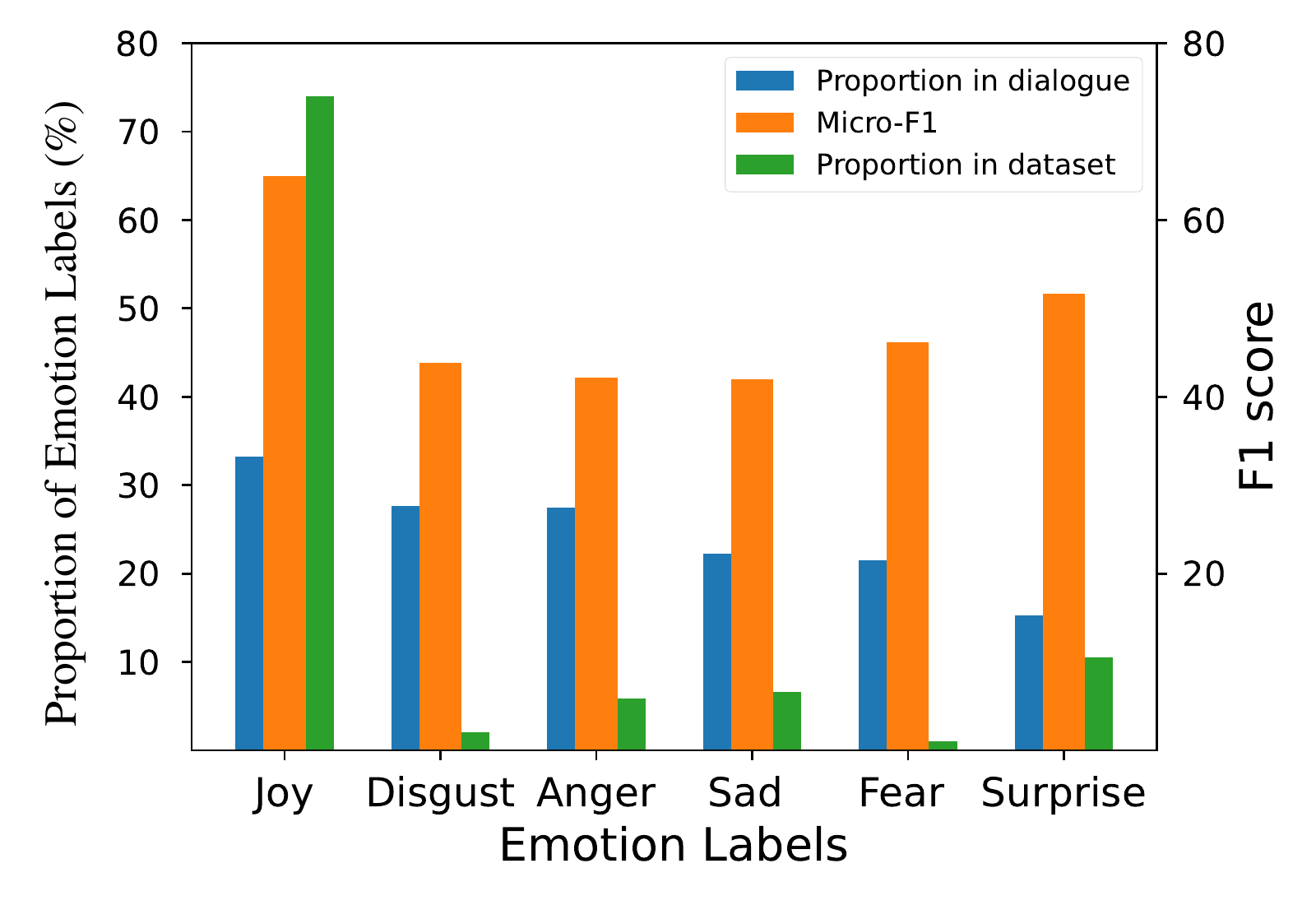}
	\caption{Effect of the class imbalance in the dialogue and that in the dataset on the performance. Note that we explore on the DailyDialog (- \texttt{Neutral}) dataset since it has an extreme class imbalance problem.}
	\label{fig:classimbalance}	
\end{figure}

We count the average proportion of each emotion in the dialogue, the proportion of each emotion in the dataset, and the corresponding F1 scores from the DailyDialog dataset except for \texttt{Neutral}, as shown in Figure~\ref{fig:classimbalance}. It can be seen that although the proportions of \texttt{Sad} and \texttt{Fear} are quite different in the dataset, they have roughly the same proportion of emotions in the conversation, resulting in similar F1 scores. The same goes for \texttt{Disgust} and \texttt{anger}. However, the F1 score of \texttt{suprise} with the lowest proportion in dialogue is higher than that of other four emotions except for \texttt{Joy}, which can be attributed to the fact that the high proportion in the dataset also plays a role in the training process. Note that \texttt{Joy} has a large proportion in both the dataset and the dialogue. Although the proportion of \texttt{Joy} in the dataset is several times higher than those of other emotions, it has only a slightly higher F1 score than \texttt{Surprise}, which confirms that the proportion in the dialogue is more critical to the model performance than the proportion in the dataset.

\subsection{Comparison of modeling emotional inertia and contagion}
In this subsection, we discuss the discrepancy in modeling based on between emotional inertia and contagion in the proposed model. First, we extract data samples with emotional inertia and contagion in the test set. Then, we test these samples by employing F1 scores, i.e., as evaluation indicators of emotional inertia and contagion. Note that in the training or testing phase, we use all the samples in the training or test set; whereas in the calculation of F1 scores, only the samples related to emotional inertia or contagion are employed. The data samples with emotional inertia and contagion are extracted in the following ways. 
\\ \noindent 
\textbf{(1)} Emotional inertia samples. If the emotion $y_{t}^{p_i}$ of participant $p_i$'s utterance $u_{t}^{p_i}$ at moment $t$ is the same as the emotion $y_{\bm{s}(t)}^{p_i}$ of $p_i$'s utterance $u_{\bm{s}(t)}^{p_i}$ at the previous moment $\bm{s}(t)$ (i.e., $y_{t}^{p_i}=y_{\bm{s}(t)}^{p_i}$), then $u_{t}^{p_i}$ and $u_{\bm{s}(t)}^{p_i}$ are taken as emotional inertia samples. 
\\ \noindent 
\textbf{(2)} Emotional contagion samples. If the emotion $y_{t}^{p_i}$ of participant $p_i$'s utterance $u_{t}^{p_i}$ at moment $t$ is the same as the emotion $y_{\bm{o}(t)}^{p_j}$ of interlocutor $p_j$'s utterance $u_{\bm{o}(t)}^{p_j}$ at the moment $\bm{o}(t)$ (i.e., $y_{t}^{p_i}=y_{\bm{o}(t)}^{p_j}$), then $u_{t}^{p_i}$ and $u_{\bm{o}(t)}^{p_j}$ are taken as emotional contagion samples. 

\begin{table}[htbp]
\caption{F1 scores of EmotionIC on the samples with emotional inertia and contagion for different datasets. We adopt micro F1 score to record the results on the DailyDialog dataset and adopt weighted F1 score on the other datasets.}
\label{tab:InerandCont}	
\begin{center}
\renewcommand{\arraystretch}{1.0}
\setlength{\tabcolsep}{10pt}
\begin{tabular}{c|c|cccc}
\toprule
\multicolumn{2}{c|}{Datasets} &  IEMOCAP & DailyDialog & MELD & EmoryNLP\\ 
\cline{2-5}
\hline
\multirow{2}{*}{Emotional Inertia} &\#Samples &1,151 & 454 & 861 &242\\
							& F1&77.12 &74.08 &72.59 &47.04\\
\hline
\multirow{2}{*}{Emotional Contagion} &\#Samples &410 &670 &1,003 &497\\
							& F1&49.82 &73.60 &61.07 &31.66\\
\bottomrule
\end{tabular}
\end{center}
\end{table}
Table~\ref{tab:InerandCont} shows that the modeling ability of our EmotionIC in terms of emotional inertia is superior to that in terms of emotional contagion. It is intuitive that the ability of the ERC model to simulate emotional inertia and contagion is to some extent related to the number of corresponding samples. That is, the larger the sample size of emotional inertia or contagion, the better the model should perform on the corresponding samples. However, the results in Table~\ref{tab:InerandCont} suggest that it is not the case. Except for the IEMOCAP dataset, all other datasets show the opposite results. In other words, although the sample size of emotional contagion is larger than that of emotional inertia, the corresponding F1 scores are still lower, suggesting that utterances involving emotional contagion in the conversation are more difficult to classify accurately.

\begin{table}[htbp]
\caption{F1 scores of EmotionIC on the samples with emotional inertia and contagion for different emotions. These results are derived from the DailyDialog (- \texttt{Neutral}) dataset.}
\label{tab:InerandCont_DD}	
\begin{center}
\renewcommand{\arraystretch}{1.0}
\setlength{\tabcolsep}{8.5pt}
\begin{tabular}{c|c|cccccc}
\toprule
\multicolumn{2}{c|}{Emotions} &  Joy & Anger & Sadness & Fear & Surprise & Disgust\\ 
\cline{2-5}
\hline
\multirow{2}{*}{Emotional Inertia} &\#Samples &330 & 60 & 31 &5 &8 &20\\
							& F1&79.05 &63.27 &55.81 &75.00 &62.50 &57.14 \\
\hline
\multirow{2}{*}{Emotional Contagion} &\#Samples &483 &34 &46 &9 &86 &12\\
							& F1&80.54 &42.55 &57.58 &18.18 &66.67 &35.29\\
\bottomrule
\end{tabular}
\end{center}
\end{table}
We further count the number of samples with emotional inertia and contagion for different emotions in the DailyDialog dataset, as shown in Table~\ref{tab:InerandCont_DD}. EmotionIC performs well on the samples involving emotional inertia. However, the results on the samples involving emotional contagion show a large variance due to the extreme class imbalance. This phenomenon is consistent with the findings obtained from Table~\ref{tab:InerandCont}, i.e., EmotionIC is more prone to accurately classify utterances involving emotional inertia.

\section{Conclusion}\label{sec:conclusion}
Our proposed EmotionIC is a novel approach driven by emotional inertia and contagion for the ERC task. EmotionIC adequately models a conversation at both the feature-extraction and classification levels, and consists of three main modules: IMMHA, DiaGRU, and SkipCRF. At the feature-extraction level, we utilize IMMHA to capture global contextual dependencies with identity information, and DiaGRU to extract speaker- and temporal-aware local contextual information. At the classification level, the designed SkipCRF is leveraged to capture complex emotional flows from higher-order neighboring utterances, which can explicitly simulate emotional propagations in the conversation. Since the optimal sequence of emotion labels can be obtained by utilizing SkipCRF, ERC does not require an additional softmax layer for classification. Extensive experimental results on the benchmark datasets confirm that the proposed EmotionIC can efficiently model contexts based on emotional inertia and contagion, which outperforms all baseline models.

Current ERC tasks confront several formidable challenges, such as: (1) the conversation length in the dataset is too short to facilitate contextual modeling; (2) the class imbalance in the dataset leads to recognition results being biased towards the majority class/emotion; and (3) the model has difficulty in distinguishing similar emotions. Therefore, we will strive to mitigate these issues in our future effort. Furthermore, in order to further strengthen the capacity of our model to emulate emotional contagion, we intend to investigate the effects of external commonsense knowledge and multimodal methods on emotional mutations in future work.

%%%%%%%%%%%%%%%%%%%%%%%%%%%%%%%%%%%%%%%%%%%%%%%%%%%%%%%
%%% Acknowledgements. 致谢
%%%%%%%%%%%%%%%%%%%%%%%%%%%%%%%%%%%%%%%%%%%%%%%%%%%%%%%

\hspace*{\fill}

\noindent
\textbf{Acknowledgements} \ This work was supported in part by the National Natural Science Foundation of China under Grant 62236005, 61936004, and U1913602. We express our thanks to the people who helped for this work, and acknowledge valuable suggestions from the reviewers.
% \Acknowledgements{This work was supported in part by the National Natural Science Foundation of China under Grant 62236005, 61936004 and U1913602.}

%%%%%%%%%%%%%%%%%%%%%%%%%%%%%%%%%%%%%%%%%%%%%%%%%%%%%%%
%%% Supplements. 补充材料, 非必选
%%%%%%%%%%%%%%%%%%%%%%%%%%%%%%%%%%%%%%%%%%%%%%%%%%%%%%%
% \Supplements{Appendix A.}

%%%%%%%%%%%%%%%%%%%%%%%%%%%%%%%%%%%%%%%%%%%%%%%%%%%%%%%
%%% Reference section. 参考文献
%%% citation in the content using "some words~\cite{1,2}".
%%% ~ is needed to make the reference number is on the same line with the word before it.
%%%%%%%%%%%%%%%%%%%%%%%%%%%%%%%%%%%%%%%%%%%%%%%%%%%%%%%
% \begin{thebibliography}{99}

% \bibitem{1} Author A, Author B, Author C. Reference title. Journal, Year, Vol: Number or pages

% \bibitem{2} Author A, Author B, Author C, et al. Reference title. In: Proceedings of Conference, Place, Year. Number or pages

% \end{thebibliography}
% \small
% \footnotesize
\scriptsize
\bibliographystyle{gbt7714-numerical} % a modified version of IEEEtran.bst
\bibliography{emotionic}

%%%%%%%%%%%%%%%%%%%%%%%%%%%%%%%%%%%%%%%%%%%%%%%%%%%%%%%
%%% Appendix sections. 附录章节, 非必选
%%%%%%%%%%%%%%%%%%%%%%%%%%%%%%%%%%%%%%%%%%%%%%%%%%%%%%%
%\begin{appendix}
%\section{Name}

%\end{appendix}

\end{document}